\documentclass{article}

\usepackage[preprint]{nips_2018}

\usepackage[utf8]{inputenc} % allow utf-8 input
\usepackage[T1]{fontenc}    % use 8-bit T1 fonts
\usepackage{hyperref}       % hyperlinks
\usepackage{url,color}            % simple URL typesetting
\usepackage{booktabs}       % professional-quality tables
\usepackage{amsfonts}       % blackboard math symbols
\usepackage{nicefrac}       % compact symbols for 1/2, etc.
\usepackage{microtype}      % microtypography
\usepackage{amsmath}
\usepackage{graphicx}
\usepackage{bm}
\usepackage{verbatim}
\usepackage{natbib}

\newcommand{\mycomment}[1]{}

\DeclareMathOperator*{\argmax}{arg\,max}
\title{A Dissection of Overfitting and Generalization in Continuous Reinforcement Learning}

\author{
  Amy Zhang \\
  McGill University \\
  Facebook AI Research \\
  \texttt{amyzhang@fb.com} \\
  \And
  Nicolas Ballas \\
  Facebook AI Research \\
  \texttt{ballasn@fb.com} \\
  \AND
  Joelle Pineau \\
  McGill University \\
  Facebook AI Research \\
  \texttt{jpineau@fb.com}
}

\begin{document}
% \nipsfinalcopy is no longer used

\maketitle

\begin{abstract}
The risks and perils of overfitting in machine learning are well known. However most of the treatment of this, including diagnostic tools and remedies, was developed for the supervised learning case. In this work, we aim to offer new perspectives on the characterization and prevention of overfitting in deep Reinforcement Learning (RL) methods, with a particular focus on continuous domains.  We examine several aspects, such as how to define and diagnose overfitting in MDPs, and how to reduce risks by injecting sufficient training diversity.  This work complements recent findings on the brittleness of deep RL methods and offers practical observations for RL researchers and practitioners.
\end{abstract}

\section{Introduction}
% The risks and perils of overfitting in machine learning are well known~\cite{domingos12}. However most of the discussion of this phenomenon, including diagnostic tools and remedies, were developed for the supervised learning case.  There is relatively little discussion of overfitting, and the related notion of generalization, in the context of unsupervised or reinforcement learning.   The very notion of overfitting is somewhat problematic in reinforcement learning, where it is not possible to properly estimate generalization error on a hold-out test set (due to possible shift in the data distribution induced by the learned policy).

Machine learning is concerned with learning a function from data.  That data typically consists of a mix of \textit{pattern} and \textit{noise}.   For most supervised learning applications, the data comes from the natural world, where the pattern may be (near) infinitely complex, and the true noise is not known.
Yet recent studies on generalization in \textbf{supervised learning} have shown that deep neural networks, with their overparameterization and massive capacity, can lead to memorization of the data, and thus poor generalization \citep{2018arXiv180406893Z, arpit17}. 

One of the objectives of our work is to explore if similar issues occur in \textbf{reinforcement learning} (RL).
Indeed, recent findings have shown significant brittleness of deep RL methods~\citep{henderson18}.
One of the challenges with RL, is that almost all research is done using simulators.  By design, these simulators consist of finite and stationary lines of code, running a deterministic program, where the only injection of \textit{noise} is through an initial random seed.  In this case, it is not surprising that (for a fixed random seed), given sufficient data, the RL method will exhaust the noise.  In effect, memorization is inevitable. 
This is especially prevalent in simple tasks, those with small state spaces, simple perception, short planning horizon, and deterministic transition functions.  This constitutes a large portion of benchmark domains considered in the field~\citep{1606.01540,machado17arcade,DBLP:journals/corr/BeattieLTWWKLGV16}.

In this work we examine conditions under which deep RL achieves generalization vs memorization, with both model-free and model-based reinforcement learning methods.
We investigate the prevalence of memorization with randomization experiments in both the discrete and continuous action setting, as well as with value-based and policy-based methods.
% We also explore model-based RL, to see if learning a dynamics model leads to more or less brittle generalization abilities. 
Through randomized reward experiments, we develop strategies to test memorization. We also propose tests to investigate robustness of the learned policy specific to the continuous domain, by expanding the initial state distribution and adding stochastic noise to the observation space.

%a variety of environments with continuous state space $\mathcal{S}$
%For these environments we use Double DQN \citep{DBLP:journals/corr/HasseltGS15}, with prioritized experience replay \citep{DBLP:journals/corr/SchaulQAS15}.
%We then analyze the continuous action space setting with MuJoCo \citep{conf/iros/TodorovET12} environments (Reacher and Thrower). % using Proximal Policy Optimization (PPO) \citep{DBLP:journals/corr/SchulmanWDRK17}
We first explore classic RL simulated environments such as the Cartpole and Acrobot control tasks in the Gym toolkit \citep{1606.01540} with discrete action space and also look at the continuous action space setting with MuJoCo Reacher and Thrower environments \citep{conf/iros/TodorovET12}. % using Proximal Policy Optimization (PPO) \citep{DBLP:journals/corr/SchulmanWDRK17}
As the goal of RL is eventually to transfer to to the real world, 
we propose new reinforcement learning tasks that received observations from natural images and explore generalization in that setting as well.

%% FIXME Work in progress
Our analysis shows that deep RL algorithms can overfit both in standard simulated environment and when operating on natural images. However, as soon as there is enough training data diversity in the simulated environment, deep RL generalizes well. It  is likely due to the fact that these  commonly used synthetic domains have relatively little noise and spurious patterns. On the other hand, deep RL algorithms show more prominent overfitting when observing natural data. Those observations advocate for the development of new benchmarks in order to study more thorough overfitting in deep RL.

%FIXME %increased tendencies for deep neural networks to overfit in high-dimensional observation spaces such as the pixel domain. However, the overfiting remains limited in those synthetic tasks due FIXME.

%We will then give a brief introduction to the reinforcement learning setting in Section \ref{sec:rl_bg}, and visit the idea of generalization in supervised learning and explore work in that area in Section \ref{sec:sl_gen}. Then we re-define generalization in this new framework in Section \ref{sec:rl_gen}. 

%What's the gap for RL? 
%Review formal definitions of generalization from RL literature (e.g. S. Murphy's JMLR paper) extend to notions of transfer and zero-shot learning
%Introduce and discuss the train/test seeds approach
%Explain overfitting in RL (what it means, how to detect, how to avoid)
%Describe memorization in supervised learning
%Propose new method for checking memorization in RL
%Anything specific to continuous domains?
%Add lots of references.
\section{Technical Background}
\label{sec:rl_bg}

\textbf{Reinforcement Learning.}
We examine the setting where the environment is a Markov Decision Process (MDP), represented by a tuple $M=(\mathcal{S}, \mathcal{A}, T, R, \mathcal{S}_0, \gamma)$, where $\mathcal{S}$ is a state space, $\mathcal{A}$ an action space,  $T(s, a, s')$ is the state-to-state transition distribution with $s, s'\in \mathcal{S}$, $R(s,a)$ is the reward function, $\mathcal{S}_0$ the initial state distribution, $\gamma\in[0,1)$ a discount factor~\citep{bellman57}. The policy $\pi: \mathcal{S} \rightarrow \mathcal{A}$ defines an action selection strategy.  Reinforcement learning (RL) is concerned with finding an optimal policy, denoted $\pi^*: \mathcal{S} \rightarrow \mathcal{A}$, that maximizes the expected sum of rewards.

\textbf{Value function estimation.}
The value function defines the expected sum of rewards over the trajectory $t=0 ... T$ while following policy $\pi$:
$V^\pi(s)= \mathbb{E}[\sum_{t=0}^T\gamma r_t |s_0=s]$.
The state-action Q-function is similarly defined and can be recursively estimated from Bellman's  equation:
$Q^\pi(s_t,a_t)= r_t + \gamma \sum_{s' \in S} T(s,a,s') max_{a' \in A} Q(s',a').$

In the absence of a known transition model $T()$, the Q-function can be estimated from a batch of data using fitted Q-iteration, which learns an approximate function $\hat{Q}(\cdot,\theta)$, parameterized by $\theta$, minimizing the loss:
$\mathcal{L}(\theta) = \left( 
\hat{Q}(s_t,a_t,\theta) - Y_t  \right)^2,
\textrm{ where } Y_t = r_t + \gamma \max_a Q(s_{t+1},a,\theta).$

\textbf{Deep RL.} 
The Deep Q-Network (DQN) extends fitted Q-iteration by using a (potentially large) neural network for the Q-function~\citep{mnih-dqn-2015}.
 Several additional steps are helpful in stabilizing learning, including using a target network from a previous iteration, applying L1-smoothing of the loss, incorporating experience replay~\citep{mnih-dqn-2015}, and correcting with Double DQN~\citep{DBLP:journals/corr/HasseltGS15}. This class of approach is suitable for continuous state, discrete action MDPs.

\textbf{Policy-based Methods.}
In the case of continuous action domains, policy-based methods are typically used. Here we directly estimate a parameterized form of the optimal policy $\pi_\theta(s)$, where in deep RL $\pi_\theta(s)$ takes the form of a neural network. This network takes as input the state $s_t$, outputs a distribution over possible actions $a_t$, and is trained with a loss of the form:
$\mathcal{L}_{\text{PG}}(\theta)=\hat{\mathbb{E}}_t\big[\log\pi_\theta(a_t|s_t)\hat{A}^\pi(s_t, a_t)\big]$.
%The advantage function is A^\pi_t:=Q^\pi(s_t,a_t)-V^\pi(s_t)$, and \hat{A}^\pi$ is an estimator. We learn an approximation of $V^\pi$, as with actor-critic, and use this as our surrogate objective with additional clipping to reduce variance further. 
Proximal Policy Optimization (PPO) \citep{DBLP:journals/corr/SchulmanWDRK17} is a popular variant of policy gradient methods that  incorporates constraints on the policy change, as well as mini-batches and multiple epochs of learning between each query of the environment. 
% This brings this closer to supervised learning, and we do see an increase in overfitting when increasing the number of epochs.

%\nicolas{Overall this section is very quicly specific and from the text it is not clear why we are describing thoses models. It think we can add this at the end of the methodoly section in a Experimental setting part? }

\section{Perspectives on generalization and overfitting}
\subsection{Generalization in Supervised Learning}
\label{sec:sl_gen} 
In Supervised Learning (SL), we assume an unknown data distribution $\mathcal{D}$ from which we draw i.i.d samples $\mathcal{X}_{\text{train}}$ with ground truth labels $\mathcal{Y}_{\text{train}}$ where $y\in\mathcal{Y}$ are extracted from a target concept $y=c(x)$. The goal is to learn a function $f$ that learns the mapping from $\mathcal{X}_{\text{train}}$ to $\mathcal{Y}_{\text{train}}$. Given a loss function $l(f(x), y)$, the generalization error $E_G$ is defined as the difference in true and expected error. 
\begin{equation}
E_G = E_{x,y\sim\mathcal{D}}[l(f(x), y] - \frac{1}{N}\sum_{i=1}^N l(f(x_{tr,i}),y_{tr, i})
\end{equation}
In practice, however, we do not have access to underlying joint probability distribution $p(x,y)$. Instead, we compute a proxy empirical error using a withheld test set $(\mathcal{X}_{\text{test}}, \mathcal{Y}_{\text{test}})$ drawn from the same distribution as $(\mathcal{X}_{\text{train}},\mathcal{Y}_{\text{train}})$. This gives us the following formula for generalization error: 
\begin{equation}
E_G = \frac{1}{M}\sum_{i=1}^M l(f(x_{te,i}),y_{te,i}) - \frac{1}{N}\sum_{i=1}^N l(f(x_{tr,i}),y_{tr, i})
\end{equation}
where $N=|\mathcal{X}_{\text{train}}|, M=|\mathcal{X}_{\text{test}}|, (x_{tr,i},y_{tr,i})\in(\mathcal{X}_{\text{train}}, \mathcal{Y}_{\text{train}})$, and $(x_{te,i},y_{te,i})\in(\mathcal{X}_{\text{test}}, \mathcal{Y}_{\text{test}})$.
\textbf{Overfitting} is defined as poor generalization behavior, or large $E_G$.

Recent work \citep{2016arXiv161103530Z} explores the issue of generalization performance in the supervised setting, showing that deep neural networks are capable of memorizing random data, thus exhibiting extreme overfitting.

An interesting question is whether analogous concepts of generalization, memorization and overfitting arise in the reinforcement learning framework.

%A perhaps related question, which we do not tackle in this paper, is the notion of overfitting in unsupervised data. Comment more on this?

% \nicolas{Maybe we can have a more high-level description of generalization (i.e. we want to minimize the excepted risks, but we don't have access to it so we minimize a proxy the empirical risk... }

\subsection{Generalization and Memorization in Reinforcement Learning}
\label{sec:rl_gen}

We focus on characterizing and diagnosing overfitting in deep Reinforcement Learning (RL) methods, with a specific focus on continuous domains.  The continuous state space case is particularly interesting with deep reinforcement learning because it is a setting where we expect function approximation (and particularly, a function represented by a deep neural network) to perform well.
The continuous state setting is also interesting because we cannot feasibly explore all possible initial states during training time, thus we suppose that generalization is necessary to solve the task\footnote{Our work was done concurrently with~\cite{2018arXiv180406893Z}, which focuses on the discrete state setting.}.

Considering the fact that almost all research in RL is done using simulators, to further examine generalization and overfitting it is necessary to consider cases where (1) the dataset is much smaller than the complexity of the simulator, (2) the random seed is varied to create noise (thus inducing variance in the trajectories due to different initial states or transitions), or alternately (3) the simulator itself is drawn from a random distribution (e.g. by varying the initial state distribution, or transition function, or reward function).   We investigate all three cases in experiments below.  Cases 1 \& 2 correspond to within-task generalization; case 3 corresponds to out-of-task (or transfer) generalization.

%However, in the more interesting case, there are differences in the train and test environments $M_\text{train}$ and $M_\text{test}$, or the state space of the environment is large and varied enough that it is intractable to traverse all of it during training. In these types of tasks, we must rely on the model to generalize well. However, a few key components in SL for defining generalization are missing in the RL paradigm. In most problems, we do not have access to $\mathcal{Y}_{\text{train}}$. RL also lacks the same notion of a withheld evaluation set of states or tasks. 

In the within-task case, generalization is achieved when a policy trained on an initial set of training trajectories is shown to perform well on a set of test trajectories generated by the same simulator.  In the out-of-task case, generalization is achieved when a policy trained on one setting of the simulator performs well on another setting of the simulator.

As exposed above, the only source of noise when doing RL in a simulated environment is through the random seed.  Thus, as also proposed in ~\cite{2018arXiv180406893Z}, the only way to achieve a separation between \textit{training} and \textit{testing} sets is by separation of random seeds.  Define $\mathcal{S}_{0, train}$ to be the trajectories generated by the set of training seeds, and $\mathcal{S}_{0, test}$ to be the set of trajectories generated by the testing seed.

We can now define generalization in RL:
\begin{equation}
E_{G,RL} = \frac{1}{N}\sum_{i=1}^N \sum_{t=1}^T R(s_t, \pi_\theta(s_t)|s_0=s_{tr,i}) - \frac{1}{M} \sum_{i=1}^M\sum_{t=1}^T R(s_t,\pi_\theta(s_t)|s_0=s_{te,i}),
\end{equation}
where $N$ is the number of training seeds, $M$ is the number of test seeds, $s_{tr,i}\in\mathcal{S}_{0, train},s_{te,i}\in\mathcal{S}_{0, test}$, $T$ is the length of each episode, and $\pi_\theta$ is our parameterized policy in the policy-based method case, and $\pi_\theta=\argmax_a Q_\theta(s, a)$ in the value-based method case.  In practice, for all of our experiments, we fix the number of test seeds to be $M=100$.
% \nicolas{It would be nice to motivate the different proposal, i.e. we propose train and test seeds as a method to measure overfitting in RL, Randomized reward as a way to detect memorization (need to define memorization as well) ... Also I don't know if it make sense, but maybe we can formalized the train/test seeds, randomized rewards using the RL framework to make our proposal more concrete?}
% \nicolas{Also I think it might be worth it to introduce the notion of distribution on initial seeds here}

%We further define memorization as the learning of random patterns which have no hope of generalization.

In addition to simulated environments, we also propose new tasks where observations are grounded in natural data. 
We investigate deep-RL generalization behaviors on tasks where noise comes from a natural source in that setting.

\section{Overfitting and memorization in the within-task case}
%In supervised learning, we do not have to define the boundaries of the task. There is an existing natural distribution that we sample from, and we can usually assume that distribution is fixed across training and evaluation. 

%In reinforcement learning, we often use a simulated environment as a proxy for a real world task. In this case, and in the case of our experiments, we do set artificial boundaries around how we define a single task. However, we often want our policy to be robust to unforeseen changes in the environment. How do we diagnose overfitting in this setting in an environment invariant way?

In this section we perform a series of empirical demonstrations to illustrate the various incarnations of overfitting in RL.  Several details of the experimental setup, including domain specification, deep learning architecture and training regime are included in the appendix.  They are not necessary to understand the main findings, but may be useful for reproducing the work.

\subsection{On the effect of the number of training random seeds}
Consider the case where we fix the number test seeds to $M=100$ and vary the number of training seeds.
We track the difference in performance on the initial state drawn from the set of train seeds vs. the set of test look at the difference in performance.  

\begin{figure}[h]
\centering
\includegraphics[trim={0cm 0cm 0cm 0},clip,width=.4\linewidth]{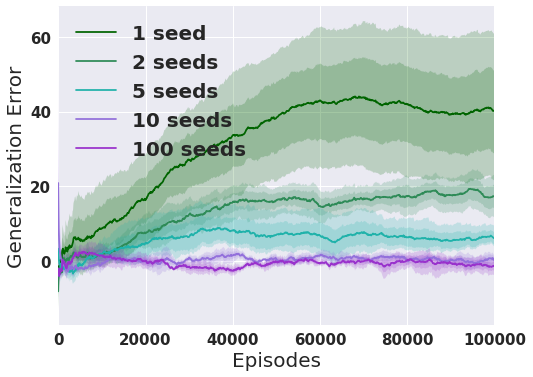}
\hspace{1.5cm}
\includegraphics[trim={0cm 0 0cm 0cm},clip,width=.4\linewidth]{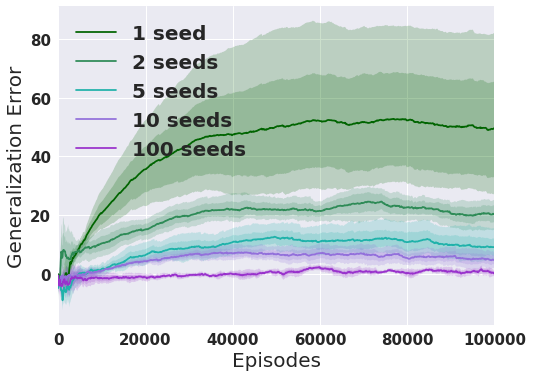}
 \caption{The 6-dim Acrobot (left) and Pixel Acrobot (right), varying number of train seeds from 1 to 100, with $\gamma=0.99$. Averaged over 5 runs. Trained for 10K episodes (6-dim) and 100K episodes (Pixel).}
\label{fig:acrobot_num_seeds}
\vskip -0.1in
\end{figure}

\textbf{Discrete Action Space.}
Experiments with discrete action spaces use a Double DQN architecture.  Full implementation details are in the appendix and code will be released.
We first consider the popular Acrobot task~\citep{Sutton96generalizationin}.  We look at 2 versions of this domain: a first where the state space is defined over the 6-dimensional joint position \& velocity space, and a second where the state is described in pixel space from a visual representation of the domain. In the pixel case
we concatenate multiple consecutive screens to allow velocity estimation, and thus circumvent partial observability.
Figure \ref{fig:acrobot_num_seeds} shows overfitting in both representations, for smaller number of training seeds\footnote{Note that with a single training seed, the initial state drawn is always the same. However, the number of states seen during training is still $|\mathcal{A}|^{|e|}$ where $|e|$ is the length of episode $e$ when using a discrete action space.}.  When the number of training seeds gets to be around 10, there is sufficient diversity in training data to generalize well given the simplicity of the environment.

\begin{figure}[h]
\centering
\includegraphics[trim={0cm 0cm 0cm 0},clip,width=.4\linewidth]{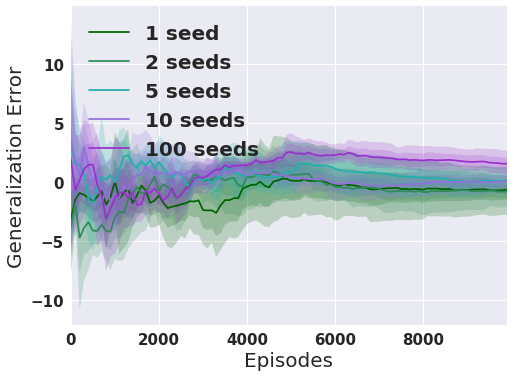}
\hspace{1.5cm}
\includegraphics[trim={0cm 0cm 0cm 0cm},clip,width=.4\linewidth]{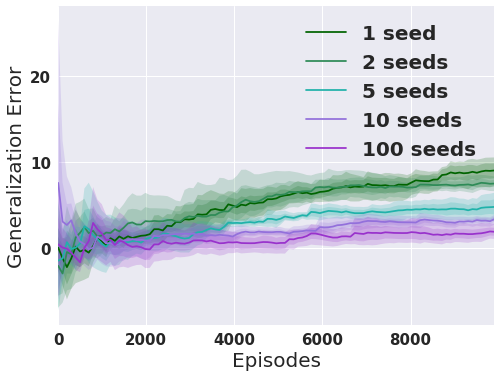}
 \caption{4-dim Cartpole (left), Pixel Cartpole (right), varying number of train seeds with $\gamma=0.995$. Averaged over 5 runs. 10K episodes.}
\label{fig:cartpole_num_seeds}
\vskip -0.1in
\end{figure}

Next we consider the ubiquitous Cartpole domain.  Again, we investigate using the standard 4-dimensional state space (position and velocity of the cart, plus angle and angular velocity of the pole). We also consider Pixel Cartpole, which uses a visual representation of the task.  For this domain, looking at Fig. \ref{fig:cartpole_num_seeds}, we do not see any overfitting with the lower dimensional representation. However overfitting is clearly seen in the Pixel Cartpole case.

In both domains, by controlling the number of seeds drawn from the initial state distribution, we can effectively control the amount of overfitting occurring.  And the number of seeds necessary is quite small, though this is likely due to the fact that these simulated domains have relatively little noise.

% \textbf{Varying $\gamma$.} Changing $\gamma$ changes the task to solve, but also has a correlation with difficulty. Low $\gamma$ correspond to lower complexity tasks, but it is difficult to evaluate this using pure reward signal, since higher $\gamma$ also corresponds to higher reward overall. 

\textbf{Continuous Action Space.}
Next we consider overfitting in continuous action spaces using PPO \citep{DBLP:journals/corr/SchulmanWDRK17} to train the RL on domains from the MuJoCo suite~\citep{conf/iros/TodorovET12} as implemented in the Gym simulator~\citep{1606.01540}.
We consider the 11-dim state space for Reacher and 23-dim state space Thrower. 
Reacher consists of two links with the task of reaching a goal point, while Thrower consists of an agent attempting to throw a ball into a goal box.  In both environments, the goal is initialized randomly for each seed (training and test). 

\begin{figure}
\centering
\includegraphics[trim={.2cm 0cm 0cm 0},clip,width=.32\linewidth]{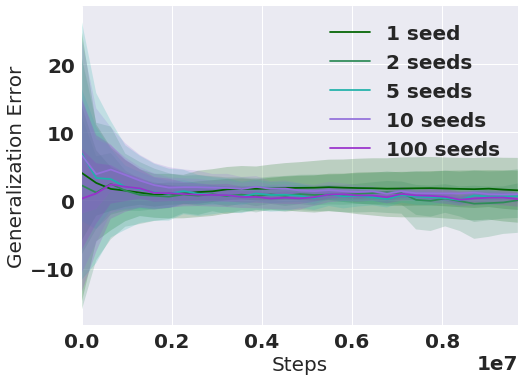}
\includegraphics[trim={0cm 0 0cm 0cm},clip,width=.32\linewidth]{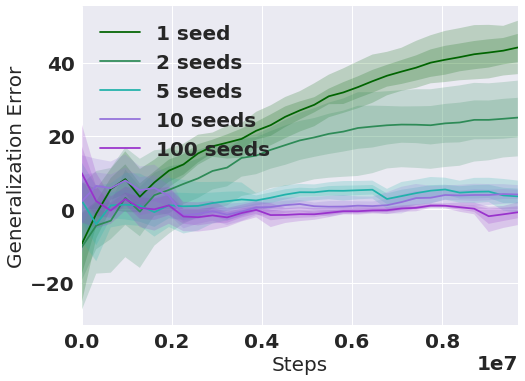}
\includegraphics[trim={0cm 0 0cm 0cm},clip,width=.32\linewidth]{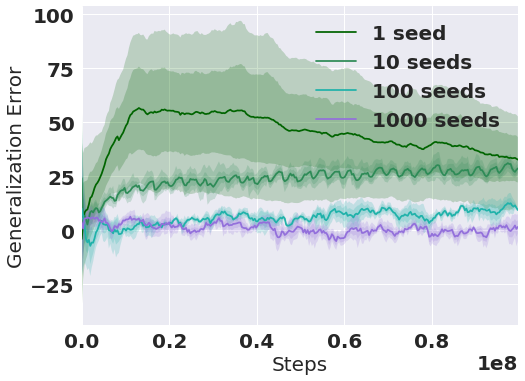}
\includegraphics[trim={0cm 0 0cm 0cm},clip,width=1\linewidth]{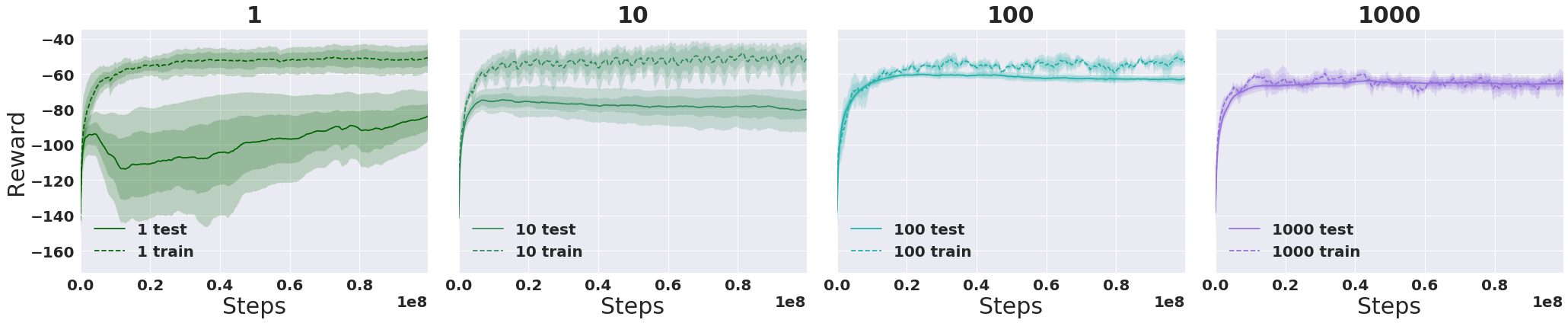}
 \caption{Reacher (top left), Thrower (top center), ThrowerMulti (top right), varying number of train seeds with $\gamma=0.99$. Averaged over 5 runs. Trained with 10M steps each. Expanded view of reward for ThrowerMulti (bottom).}
\label{fig:cont_num_seeds}
\vskip -0.1in
\end{figure}

Figure \ref{fig:cont_num_seeds} shows that the RL achieves good generalization in the Reacher, but suffers from overfitting in the Thrower with small numbers of training seeds. However at 5 training seeds and more the policy learns to generalize to unseen goal locations (as evaluated over 100 test goals).

In addition, we consider a modified Thrower environment (ThrowerMulti):  instead of a single goal box, there are now 5 goal boxes placed in random locations. The state now includes a goal id which designates which goal is the correct one -- reaching the others provides no reward.  The goal of this modification is to check whether the number of seeds necessary grows with the task complexity.
Indeed, as shown in Fig. \ref{fig:cont_num_seeds}, we need a larger number of training seeds to achieve good performance. With the addition of a single free variable, ThrowerMulti now exhibits overfitting even in the 100 seed case. We also clearly see in the 1 seed case that the policy first learns a simple policy that performs slightly better for evaluation, but over time overfits to the train case, causing evaluation performance to drop. The performance on the test seeds for ThrowerMulti is also very high variance compared to that on training seeds and compared to that on the original Thrower domain.

\begin{figure}
 \centering
\includegraphics[trim={0 0cm 0 0cm},clip,width=.31\linewidth]{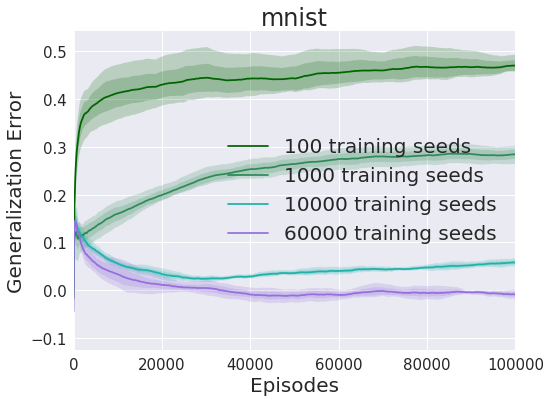}
\includegraphics[trim={0 0cm 0 0cm},clip,width=.31\linewidth]{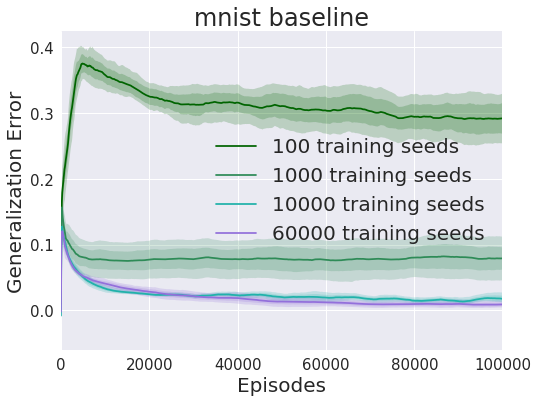}
\includegraphics[trim={0 0cm 0 0cm},clip,width=.31\linewidth]{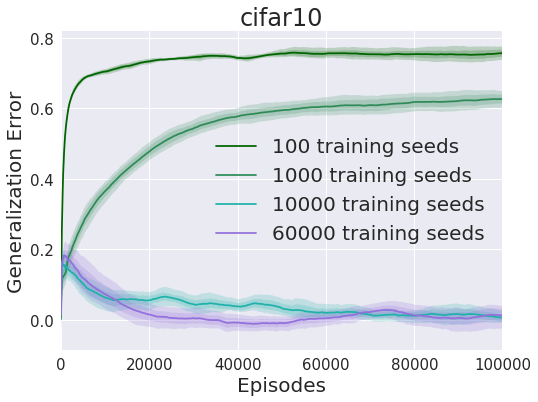}
\includegraphics[trim={0cm 0cm 0 1.2cm},clip,width=1\linewidth]{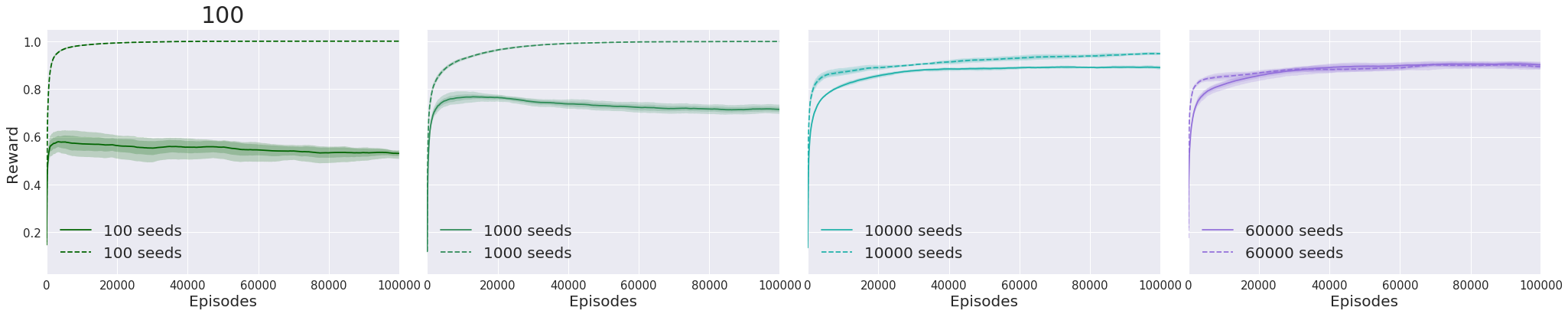}
\caption{Generalization gap for MNIST (top left) and MNIST baseline (top center) where number of seeds corresponds to number of unique images. Total reward given for MNIST (bottom). Window size $w=5$. CIFAR10 result (top right) with $w=10$.}
\label{fig:mnist}
\vskip -0.1in
\end{figure}

\textbf{Natural Images.}
We also evaluate for overfitting in natural images with a pixel-by-pixel reinforcement learning task. We use MNIST \cite{lecun-mnisthandwrittendigit-2010} and CIFAR10 \cite{cifar}, creating an exploration task where an agent is always spawned in the center of a masked image. The state consists of the agent's position and the 28x28 masked image. At each time step, the agent can move left, right, up, or down, which reveals a square window $w$ of the image. After each time step the agent classifies what the masked image is, and episode ends with reward 1 when it classifies the image successfully. Otherwise it fails after a maximum 100 time steps. We compare against a baseline where it is given the entire image to classify at the end of each episode. MNIST and CIFAR10 results are given in Figure~\ref{fig:mnist}.

With natural images, we see even in the 10k training seed case (trained on 10k images), there is a generalization gap. If we compare this with the baseline we see more overfitting effect, since there are now multiple possible partially masked images that correspond to a single image. We see similar effects in CIFAR10 results in Figure \ref{fig:mnist}.

% \subsection{Stochastic Environments}
% MuJoCo environment have deterministic transitions, so we add stochasticity by adding noise $n_s$ at training time to the state space $\mathcal{S}$. In the pixel space, this expands $\mathcal{S}$ while also injecting stochasticity in the transition function $T(s,a,s')$. We show results on pixel Acrobot.

% We also show the same effect with action noise $n_a$ for the continuous action setting with Thrower.
% \amy{TODO: show results}

\subsection{On memorization in RL: a randomized reward test}
The phenomenon of memorization in supervised learning is examined by assigning random labels to training examples~\citep{2018arXiv180406893Z,arpit17}.  This does not extend directly to the RL setting.
Instead, we propose to examine memorization by injecting randomness into the reward function.  In most continuous domains, the reward varies relatively smoothly over the state space, or at least there are few discontinuities (e.g. when reaching a goal area).
By randomizing the reward function, we can examine whether RL models exhibit memorization and learn spurious patterns that will not generalize. When there are few perturbations to the reward, or there are many training seeds, the policy should still able to learn the given reward signal. However, when the reward function becomes more erratic, it should be more difficult for the policy to train and to generalize well. 

%We also show that adding more diversity in training makes pattern recognition easier. A large amount of noise injected in the environment can be harmful in a single seed setting, but is mitigated with an increase in number of training seeds. 
% \nicolas{not sure what the conclusion of those experiments are? we can show that deep RL model memorizes but only when they trained on a very few seeds?}

To create randomized reward functions, we bin the state space into $k$ equal bins. For each bin, we uniformly at random choose a reward multiplier $\beta$ between $[-1, 1]$. These multipliers are deterministic -- they are consistent for a specific seed, but different across different initial states $s_0$. We control the amount of randomness injected into the environment with a parameter $p_{\text{rand}}$, where $p_{\text{rand}}$ is the probability of each bin having its reward signal modified by the multiplier.

\begin{figure}
\vskip -0.1in
\includegraphics[trim={0 0cm 0 14cm},clip,width=1\linewidth]{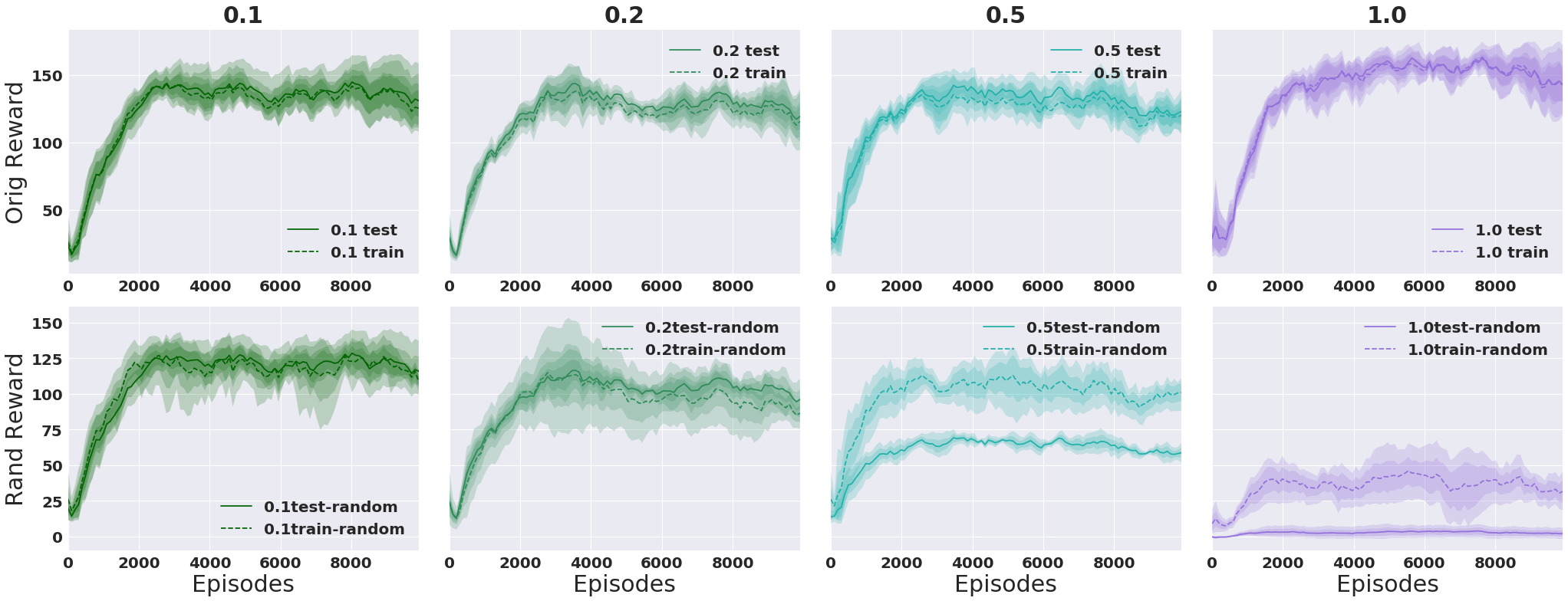}
\includegraphics[trim={0 0cm 0 14cm},clip,width=1\linewidth]{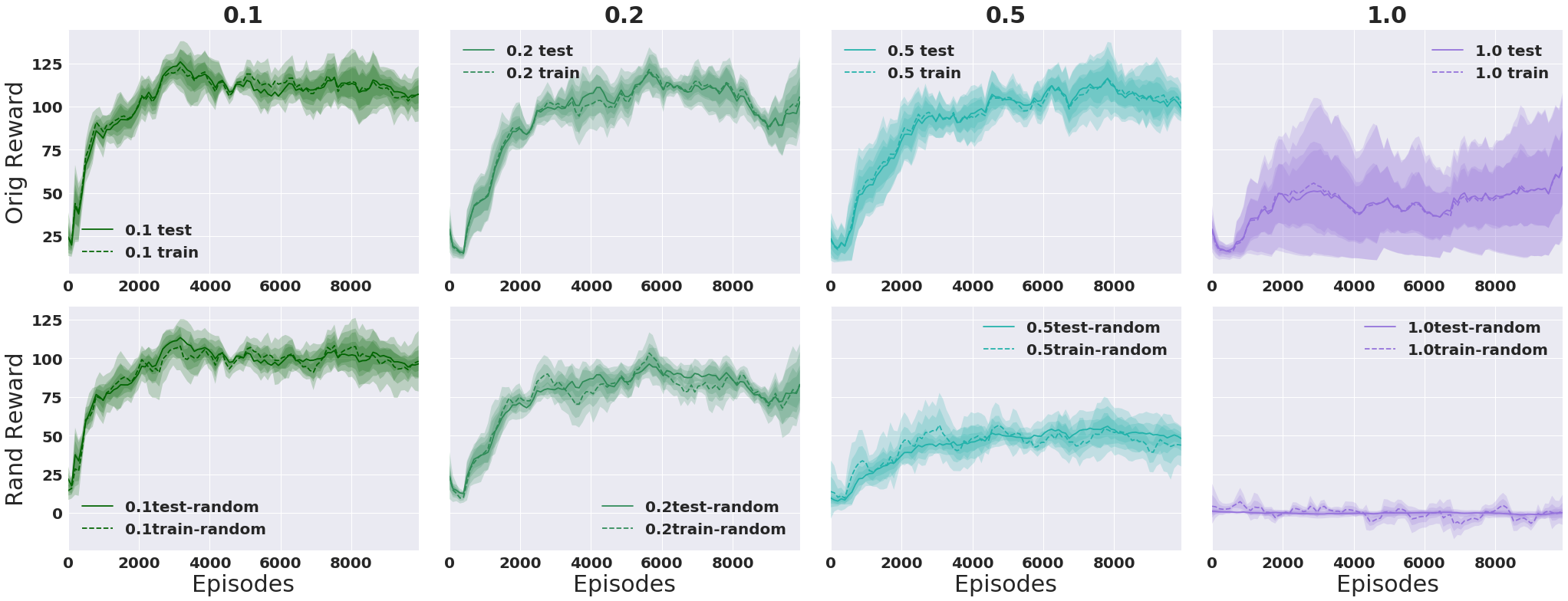}
 \caption{$k=3$. Full experiments on randomizing reward on 4-dim Cartpole: Varying $p_{\text{rand}}$ (cols) with $\gamma=0.995$ and 1 training seed (top), 100 training seeds (bottom). Averaged over 5 runs. Models are trained on the random reward and evaluated on the original reward.}
 \label{fig:rand_cartpole}
 \vskip -0.1in
\end{figure}

\textbf{Discrete Action Space.}
% \begin{figure}
% \includegraphics[trim={0 0cm 0 0cm},clip,width=1\linewidth]{figs/cartpole_rand_bin_3_num_seeds1_gamma0995_lr0003_dim512_ep10000_mem1000000.png}
%  \caption{Randomized reward with one train seed for Cartpole with $k=3$.  $\gamma=0.995$. Averaged over 5 runs. Top row shows reward for train and eval seeds on the original reward. Second row shows randomized reward for train and eval seeds. The model is trained on random reward with train seeds, with $p\in [ 0.1, 0.2, 0.5, 1]$ (columns).}
% \label{fig:rand_cartpole}
% \end{figure}
Figure \ref{fig:rand_cartpole} shows $k=3$ bins for Cartpole with a single train seed (top) and 100 train seeds (bottom). The model is trained on random reward with train seeds, with $p_{\text{rand}}\in [ 0.1, 0.2, 0.5, 1]$. In this case we see performance start to drop off as $p$ increases. For low $p_{\text{rand}}$, the model is able to effectively ignore the noise. In the 1 seed case we see the model successfully learn the random reward, especially noticeable in the $p_{\text{rand}}=0.5$ and $p_{\text{rand}}=1.0$ columns. 
However, the model does not have the capacity to memorize all the random rewards even in the 100 seed case, and so it only learns the true reward signal  up until performance collapses when $p_{\text{rand}}=1.0$, where there is no true reward signal left and the random reward is too complex to learn\footnote{Note that in the Cartpole environment any policy that can find positive reward (even random reward) and optimize for it will also perform well on the original task -- the family of policies for the original environment is a superset of policies for any reward function with a positive signal that incentivizes prolonging the episode.}.
Additional results for the Acrobot domain can be found in Appendix \ref{app:acrobot}.

\begin{figure}
\includegraphics[trim={0 0cm 0 14cm},clip,width=1\linewidth]{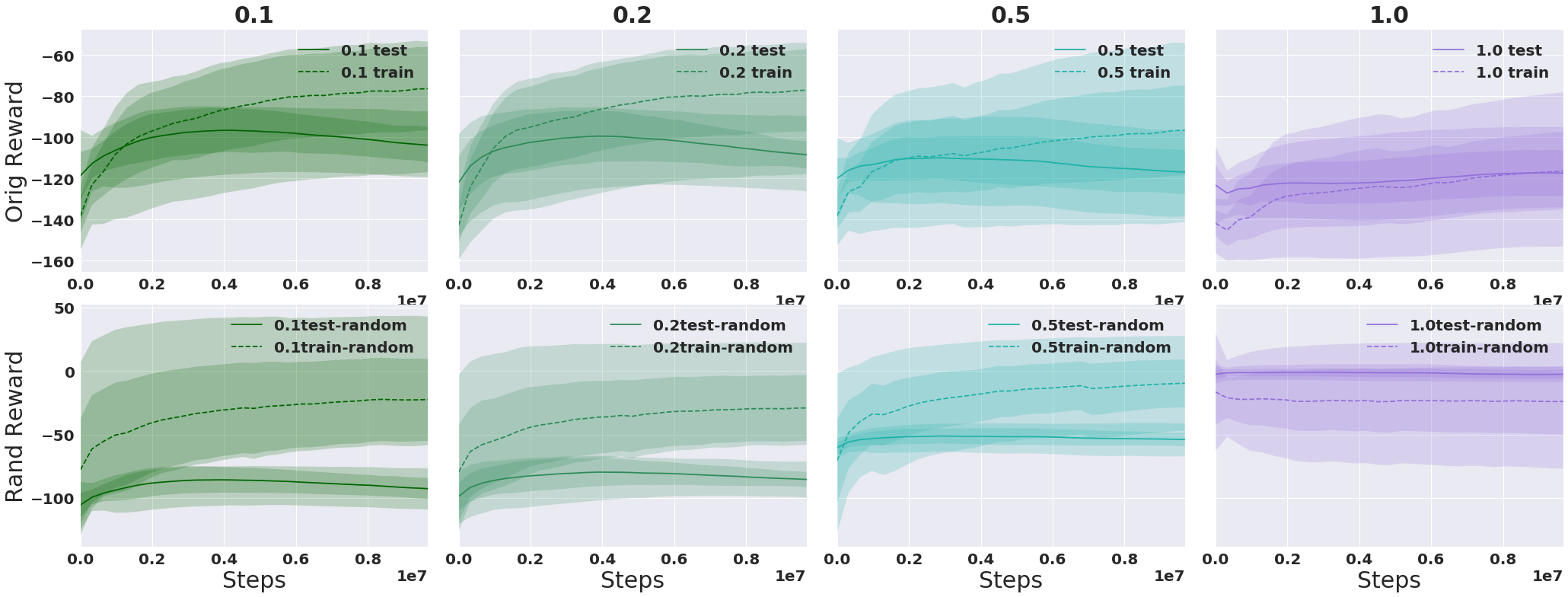}
\includegraphics[trim={0 0cm 0 14cm},clip,width=1\linewidth]{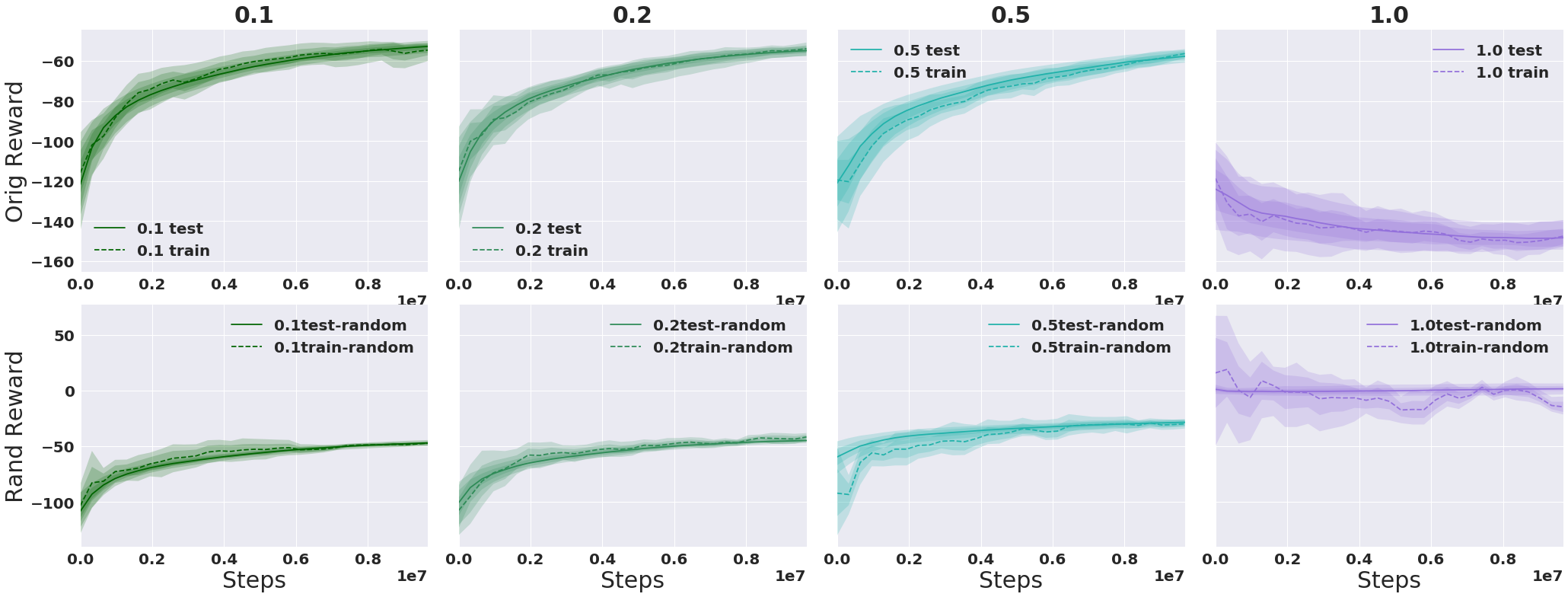}
\caption{Thrower, varying $p_{\text{rand}}$ of each bin randomized with $k=3$ and $\gamma=0.99$. Averaged over 5 runs. 1 seed (top), 100 seeds (bottom).\textsuperscript{*}}
\small\textsuperscript{*}Reward values are not directly comparable across columns because the percentage of bins modified by the random multiplier changes the average reward obtained by a random policy.
\label{fig:rand_thrower}
\vskip -0.1in
\end{figure}

\textbf{Continuous Action Space.}
Figure \ref{fig:rand_thrower} shows results for Thrower with randomized rewards. Again we see that it is easy to memorize the random reward in the 1 seed case, but much harder in the 100 seed case, and so it instead learns to concentrate on the true reward signal. We observe much larger variance in the single training seed case.  Similar results for Reacher can be found in Appendix \ref{app:reacher}.

\begin{figure}
\centering
\includegraphics[trim={0 0cm 0 0cm},clip,width=1\linewidth]{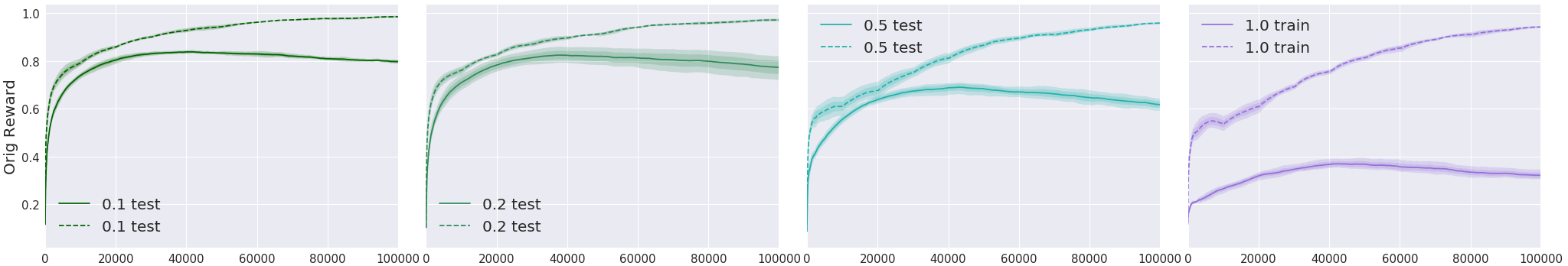}
\caption{Train and test performance with increasing $p$ from left to right on MNIST 10k. Window size $w=5$.}
\label{fig:rand_mnist}
\vskip -0.1in
\end{figure}

\textbf{Natural Images.}
Figure \ref{fig:rand_mnist} shows results for MNIST with randomized labels. As $p$ increases we see inability to generalize to the test set, but that we can still achieve perfect memorization on the training set. We had to increase the capacity of the model and use a ResNet-18~\cite{DBLP:journals/corr/HeZRS15} as backbone to achieve perfect memorization on the 10k train set.

\section{The out-of-task case for generalization in RL}
\label{sec:transfer}
Moving beyond the case where noise is induced only via the random seed, we now consider cases where it is valuable to generalize over possible distributional shifts between training and evaluation conditions.  This is closer to the case considered in previous work on overfitting in RL~\citep{5967363,2018arXiv180406893Z}, and in the broadest sense extends to notions of transfer~\citep{taylor09} and zero/few-shot learning~\citep{oh17}.

We consider the cases where the state and action sets are the same, but noise may induce changes in transition dynamics, reward function, or initial state distribution.  There is evidence (in the discrete state setting) that if the number of tasks you train on is not varied enough, the model tends to memorize the specific environments and learn them explicitly~\citep{2018arXiv180406893Z}.  When trained on enough environments, we expect generalization to allow the RL agent to navigate efficiently on new test environments drawn from the same distribution.

Recall that our goal is not to study transfer broadly, but to examine notions of overfitting and generalization in realistic conditions.  Thus we consider two mechanisms for injecting noise into the domain. First we consider an expansion of the initial state distribution, which we implement by applying a multiplier to the initial state chosen. In the MuJoCo domains, the state is initialized uniformly at random from a given range for each dimension. We want to evaluate a specific shift where we train with initial state distribution $\mathcal{S}_{\text{train},0}$ and evaluate with different initial state distribution $\mathcal{S}_{\text{test},0}$. We investigate the setting where $\mathcal{S}_{\text{test},0}$ is a strict superset of $\mathcal{S}_{\text{train},0}$, but this evaluation can also occur in sets with no overlap. In practice, we expand this initial state $s_0$ with a multiplier $m$ to get new initial state $s_0'=ms_0$.  We vary the magnitude of the multiplier $m$ to see the effect of increasing the noise.  The results in Table~\ref{table:transfer_thrower} confirm that the generalization improves significantly with more training seeds, but is made more difficult when there is increased noise in the simulation distribution.

\begin{table}[h]
% \vskip -0.1in
\begin{center}
\begin{small}
\begin{sc}
\begin{tabular}{lc|cccccc}
\hline
&               \# of Training  & &Initial&State&Multiplier  \\
Environment &    Seeds &     $m=1$ &       $m=5$ & 		$m=10$  & 	$m=20$ & 	$m=100$ \\
\hline
 &              1 & 	-101.09 & -95.49 &   -97.24 &-101.13 & 	-110.31 	\\
Thrower& 		2 & 	-74.59 &   -77.86 & -78.38 & -78.37 & 	-84.84 \\	
& 				5 &	-52.77 &  -53.58 & -54.67 &  -53.93 & 	-64.51\\
& 				10 & -53.67 & -53.71 & -54.10 &  -56.69 & 	-70.66 \\
&				100 &-49.68 & -49.24 & -51.41 &-50.32 &  	-58.85 \\
\hline
\end{tabular}
\end{sc}
\end{small}
\end{center}
%\vskip -0.1in
\caption{Out-of-task generalization on Thrower, reporting the estimated value averaged over 100 test seeds. Columns indicate the multiplier on the initial state.}
\vskip -0.1in
\label{table:transfer_thrower}
\end{table}

Second, we evaluate policy robustness by adding Gaussian noise $n\sim\mathcal{N}(0, \sigma^2)$ directly to the observation space. We add zero-mean Gaussian noise with variance $\sigma^2$ to each observed state at evaluation, and measure robustness of the policy as we increase $\sigma^2$. The results in Table~\ref{table:transfer_noise_thrower} follow a similar trend as those in Table~\ref{table:transfer_thrower}.

\begin{table}[h]
% \vskip -0.1in
\begin{center}
\begin{small}
\begin{sc}
\begin{tabular}{lc|cccccc}
\hline
&               \# of Training  & & Variance of Noise & \\
Environment &    Seeds & $\sigma^2=0.$ & $\sigma^2=0.1$ & $\sigma^2=0.2$ \\
\hline
 &              1 & 	-99.1 & -106.0 & -114.8 \\
Thrower & 		2 & 	-74.4 &  -84.8 &  -100.0\\	
& 				5 &	    -52.3 &  -69.0 & -106.3 \\
& 				10 &    -52.7 &  -72.6 & -103.0 \\
&				100 &   -49.8 &  -64.2 &  -93.7\\
\hline
\end{tabular}
\end{sc}
\end{small}
\end{center}
%\vskip -0.1in
\caption{Transfer experiments on Thrower. Columns indicate the variance $\sigma^2$ of the Gaussian noise added to the observation space.}
\vskip -0.2in
\label{table:transfer_noise_thrower}
\end{table}

%We show that there is not a complete correlation between ranking of performance of policies on $\sigma^2=0$ and $\sigma^2>0$. We also note that the learned policy for Pixel Cartpole continues to perform poorly across settings. Looking back at the original experiment with number of seeds in Figure \ref{fig:cartpole_num_seeds} we also see that we have good train performance on a single seed, which rapidly disappears as we increase the number of seeds. Our hypothesis is that in the single seed case, the model simply memorizes the sequence of actions necessary to get a good reward, but it is incapable of doing so anymore in the case of more than one initial state. Instead of incorporating the observation, the model relies on learning a heuristic to optimize reward -- completely ignoring the input. More generalization evaluation results can be found in Appendix \ref{app:transfer}.

% However, using a model-based method has a regularizing effect on Pixel Cartpole by forcing it to take into account the observation space. We see improvement in performance from Figure \ref{fig:pixel_cartpole_num_seeds_modelbased}, and see corresponding decrease in performance when noise is added in Table \ref{table:transfer_noise_pixelcartpole}.

\section{The effect of model-based RL}
\label{sec:model_based}
\begin{figure}
\centering
\includegraphics[trim={0 0 0cm 0},clip,width=.4\linewidth]{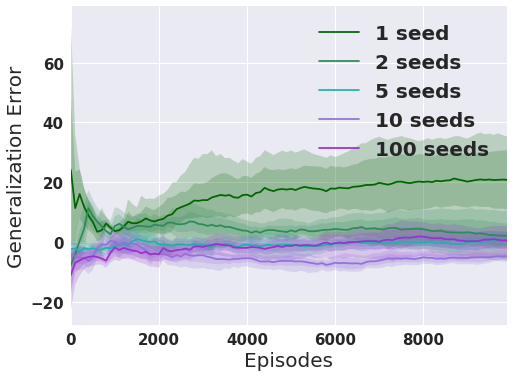}
\hspace{1.5cm}
\includegraphics[trim={0 0 0cm 0},clip,width=.4\linewidth]{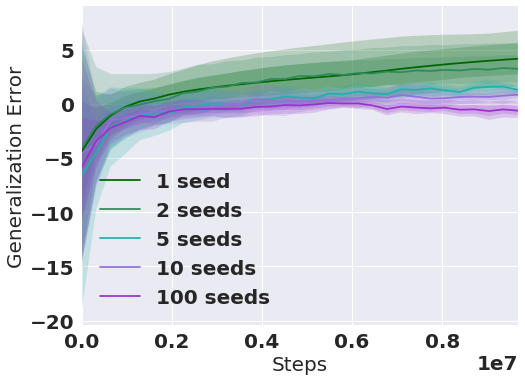}
 \caption{Model-based Reacher (left), and Thrower (right). Varying number of train seeds from 1 to 100 with $\gamma=0.99$. Averaged over 5 runs.}
\label{fig:num_seeds_modelbased}
\vskip -0.2in
\end{figure}

Several claims have been made about the fact that model-based RL improves robustness of results.  We therefore finish our investigation by looking into the generalization abilities of model-based RL methods (Fig. \ref{fig:num_seeds_modelbased}). We consider a Double DQN architecture with an additional two heads that output a prediction of next state and reward (in addition to Q-value). We consider the same modification to the critic to transfer PPO into a model-based approach.

%For Pixel Cartpole, the predicted state is in the original four-dimensional observation space, which consists of cart position, cart velocity, pole angle, and pole velocity, as opposed to predicting per pixel, which has sparsity issues. 

Results suggest that explicitly learning the dynamics model compounds existing bias in the data in the limited training seed regime. We invite the reader to examine Figure~\ref{fig:num_seeds_modelbased} and compare with the model-free results in Figure~\ref{fig:cont_num_seeds}. We see that Reacher now exhibits generalization error in the model-based case where it did not in the model-free case, and the generalization gap has increased for Thrower.
% \joelle{HOW/WHERE DO WE SEE THIS?  I commented out the Acrobot text, since I didn't the results.} \amy{added!}
% However, this depends on the complexity of the dynamics of the environment and how crucial they are to performing the task. In the case of Acrobot which consists of a two-link pendulum, which has notoriously complex dynamics, incorporating a naive model-based method where we share weights/losses of the critic with the dynamics model leads to an inability to learn a good policy (Fig~\ref{fig:num_seeds_modelbased} in comparison to model-free in Appendix~\ref{app:acrobot} Fig.~\ref{fig:acrobot_num_seeds_full_r}). 
%In other cases where dynamics are tractable and useful for solving the task, we found no significant performance gain from incorporating the model (see Figs \ref{fig:cont_num_seeds} and \ref{fig:thrower_num_seeds_modelbased}).

%  \begin{figure}
% \includegraphics[trim={0 0cm 0 0cm},clip,width=1\linewidth]{figs/Thrower-v2_random_bins_modelbased3_gamma099_lr00003_numsteps2048_procs16_numseeds1.png}
% \includegraphics[trim={0 0cm 0 0cm},clip,width=1\linewidth]{figs/Thrower-v2_random_bins_modelbased3_gamma099_lr00003_numsteps2048_procs16_numseeds100.png}
%  \caption{$b=3$. Randomized reward on Model-based Thrower: Varying $p$ (cols) with $\gamma=0.99$ and 1 training seed (top), 100 training seeds (bottom). Averaged over 5 runs.}
% \label{fig:rand_thrower_modelbased}
% \end{figure}

\section{Related Work}
\label{sec:related_work}
While not as prevalent as in the SL case, there has been some work exploring evaluation of generalization and overfitting in RL.
\cite{2017arXiv170302660R} show with linear and RBF parameterizations that they can achieve surprisingly good generalization compared to more elaborate parameterizations, and thus call for simpler linear models for better generalization. However, this is unrealistic for tasks with highly complex dynamics or nonlinear mapping between the state and value space.
\cite{2017arXiv170907796F} also looks at overfitting in the batch reinforcement learning case in POMDPs (Partially Observable Markov Decision Processes), specifically exploring the bias-variance tradeoff in the limited data scenario. They are limited to the batch setting in order to control the number of samples, but we look at the limited data setting by controlling the number of seeds, and therefore can still analyze online approaches. 
% The prescribed batch setting is also not ideal because of the confounding factor of the evaluation batch being off policy.

% classic work in generalization in RL
\cite{Murphy:2005:GEQ:1046920.1088709} analyzes generalization error for Q-learning with function approximation in the batch case.  They define generalization error as the average difference in value when using the optimal policy as compared to using the greedy policy based on the learned value function. This requires knowing  the optimal policy, which is infeasible in many environments. 
\cite{10.1007/978-3-642-11814-2_1} is a survey of abstraction and generalization in RL and also touches upon the notion of transfer as generalization. 

Concurrent work \citep{2018arXiv180406893Z} shows memorization in reinforcement learning in the discrete domain with gridworld experiments. Our analysis focuses on  continuous domains with effectively infinite number of states, which can  be more complex and closer to many real world settings.

\section{Conclusion}
Currently, much RL work is still being done in simulation, with the goal of eventually transferring methods and policies to the real world.  Of course we know that simulated environments lack the diversity -- complex signal, natural noise, nonstationary stochasticity -- of real world environments. Our natural image experiments show the discrepancy in number of seeds required for true generalization compared to the simulation environments. 
But until we drastically improve the sample complexity of RL methods, developing and testing in simulation will remain common. 
Simulated environments also continue to be a necessary tool as a controllable benchmark to compare methods, since it is not possible to ``share the real world.'' 

Given this state of affairs, we have conducted a thorough investigation into the pitfalls of  overfitting in continuous RL domains. We present a methodology for detecting overfitting, as well as evaluation metrics for within-task and out-of-task generalization. Especially in robotics tasks where it is infeasible to train on the same distribution as evaluation, our work shows the discrepancy in performance that we can expect to see in transfer scenarios.  This methodology also decouples evaluation of generalization from sample complexity for online methods. As deep reinforcement learning gains more traction and popularity, and as we increase the capacity of our models, we need rigorous methodologies and agreed upon protocols to define, detect, and combat overfitting.
 Our work contains several simple useful lessons that RL researchers and practitioners can incorporate to improve the quality and robustness of their models and methods.

%Equipped with this knowledge, we stop short of advocating for a ban on the use of simulators in RL.   

% \amy{Not sure where this goes, or if it belongs anywhere.}
% \textbf{Defining a family of tasks.} The agent's goal is to learn a policy over a family of tasks for which it does not know the distribution. This is a subset of the multi-task RL problem, in that the agent has complete information given a state observation what the task is. The mapping between the state space and reward function is 1 to 1, but the agent is not given full access to the entire state space in each episode. Therefore, we need many seeds to explore the state space sufficiently. However, there also isn't needed a 1-1 mapping of number of seeds to number of unique states, with the assumption that there is enough inherent structure in the environment that the policy should learn to generalize well enough to zero-shot transfer to an unseen state (task).

\bibliography{paper}
\bibliographystyle{plainnat}

\appendix
\newpage
\section{Model Specifications and Hyperparameters}
Convolutional head for pixel space consists of 3 convolutional layers with middle dim 512 and ReLU nonlinearities.

\subsection{Double DQN}
\begin{itemize}
    \item lr: 3e-3 for low-dim, 3e-4 for pixel
    \item target update interval: 1000
    \item replay memory: 1M for low-dim, 100K for pixel
    \item Architecture: 3-layer MLP with middle dim 512
    \item $\gamma$: 0.995 for Cartpole, 0.99 for Acrobot
\end{itemize}

\subsection{PPO}
\begin{itemize}
    \item lr: 3e-4
    \item PPO epochs: 10
    \item Number of steps between updates: 2048
    \item $\gamma$: 0.99
    \item Batch size: 32
    \item Architecture: 3-layer MLP with middle dim 512
\end{itemize}

Used Adam~\cite{DBLP:journals/corr/KingmaB14} as optimizer.
% \section{Experiment Details}

\section{Environments}
\subsection{Cartpole}
\label{app:cartpole}
\subsubsection{Random Reward}
The random reward is binning the angle $\theta$ of the pole. Each bin has a probability $p$ of having a randomized reward with its own random multiplier between -1 and 1.

\textbf{4-Dim Cartpole}
Full randomized reward experiments varying the number of training seeds, number of bins $b$, and randomization probability $p$.
\begin{figure}[h]
\includegraphics[trim={0 0cm 0 0cm},clip,width=1\linewidth]{figs/cartpole_rand_bin_3_num_seeds1_gamma0995_lr0003_dim512_ep10000_mem1000000.png}
\includegraphics[trim={0 0cm 0 0cm},clip,width=1\linewidth]{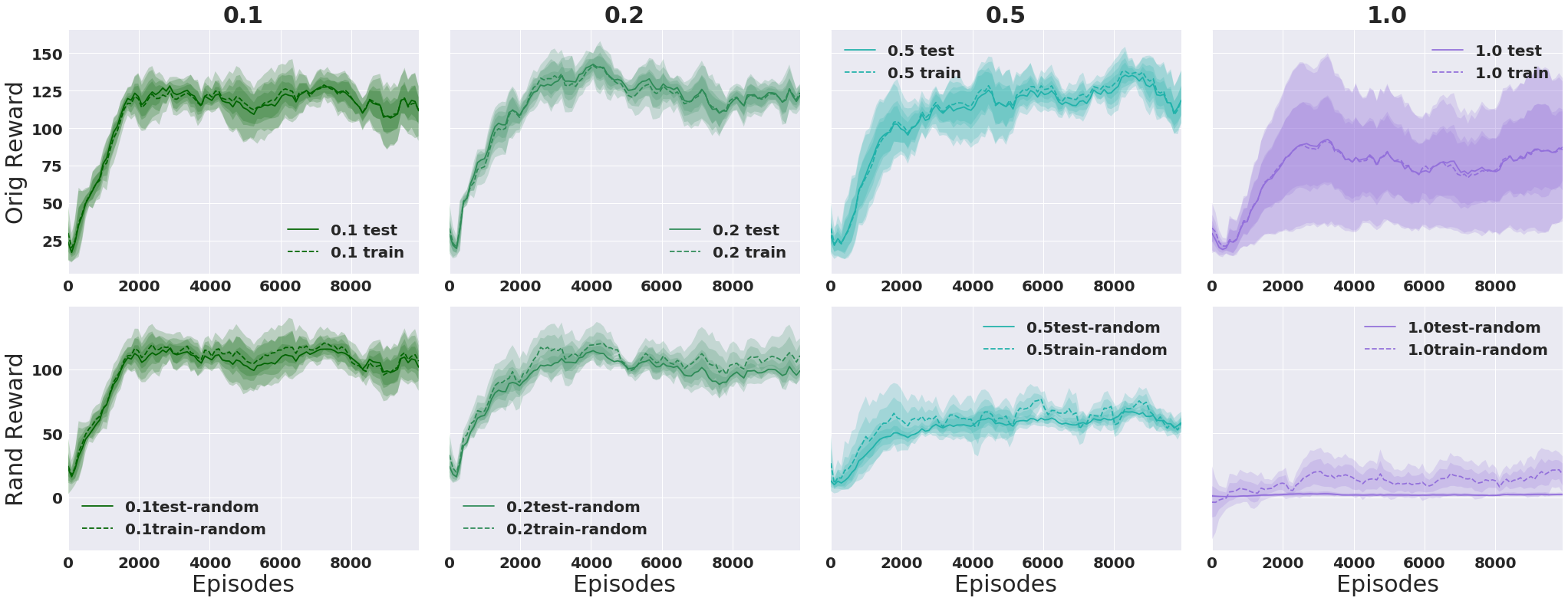}
\includegraphics[trim={0 0cm 0 0cm},clip,width=1\linewidth]{figs/cartpole_rand_bin_3_num_seeds100_gamma0995_lr0003_dim512_ep10000_mem1000000.png}
 \caption{$b=3$. Full experiments on randomizing reward on 4-dim Cartpole: Varying $p$ (cols) with $\gamma=0.995$ and varying 1, 10, 100 training seeds (rows). Averaged over 5 runs.}
 \label{fig:rand_bin3_cartpole}
\end{figure}

Figure \ref{fig:rand_bin3_cartpole} shows random rewards on Cartpole with $k=3$ bins and varying the probability $p\in [ 0.1, 0.2, 0.5, 1]$, with evaluation done with the original reward. This experiment is done on the original 4-dimensional state space for Cartpole. The original reward function of Cartpole returns a reward of 1 at every step, so we see that we still get enough positive reward signal for the original Cartpole to perform quite well even when all $k$ bins are assigned a random multiplier.

Number of episodes: 10K

\textbf{Pixel Cartpole}
Full randomized reward experiments varying the number of training seeds, number of bins $b$, and randomization probability $p$.

Number of episodes: 10K
Replay memory: 100K $(s, a, r, s')$ tuples
Learning rate: 3e-4

\begin{figure}
\centering
\includegraphics[trim={0 0 0cm 0},clip,width=.5\linewidth]{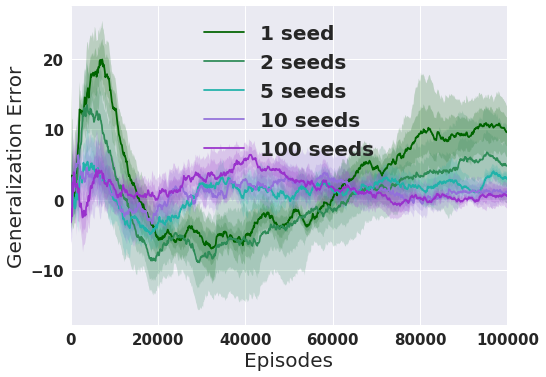}
 \caption{Model-based. Pixel Cartpole. Two states, varying number of train seeds with $\gamma=0.99$. Averaged over 5 runs.}
\label{fig:pixel_cartpole_num_seeds_modelbased}
\end{figure}

\subsection{Acrobot}
\label{app:acrobot}
Acrobot is a 2-link pendulum with only the second joint actuated. It is instantiated with both links pointing downwards. The goal is to swing the end-effector at a height at least the length of one link above the base.

The state consists of the sin() and cos() of the two rotational joint angles $\theta_1, \theta_2$ and the joint angular velocities.
\begin{equation}
[\cos(\theta_1)\sin(\theta_1)\cos(\theta_2)\sin(\theta_2) \dot\theta_1 \dot\theta_2]
\end{equation}

For the first link, an angle of 0 corresponds to the link pointing downwards.
The angle of the second link is relative to the angle of the first link.
An angle of 0 corresponds to having the same angle between the two links.
The action is either applying +1, 0 or -1 torque on the joint between the two pendulum links.
    
Number of episodes: 10K

\subsubsection{Random Reward}
We bin $\theta_0$, the angle of the first link of the pendulum. 
 \begin{figure}[h]
\includegraphics[trim={0 0cm 0 0cm},clip,width=1\linewidth]{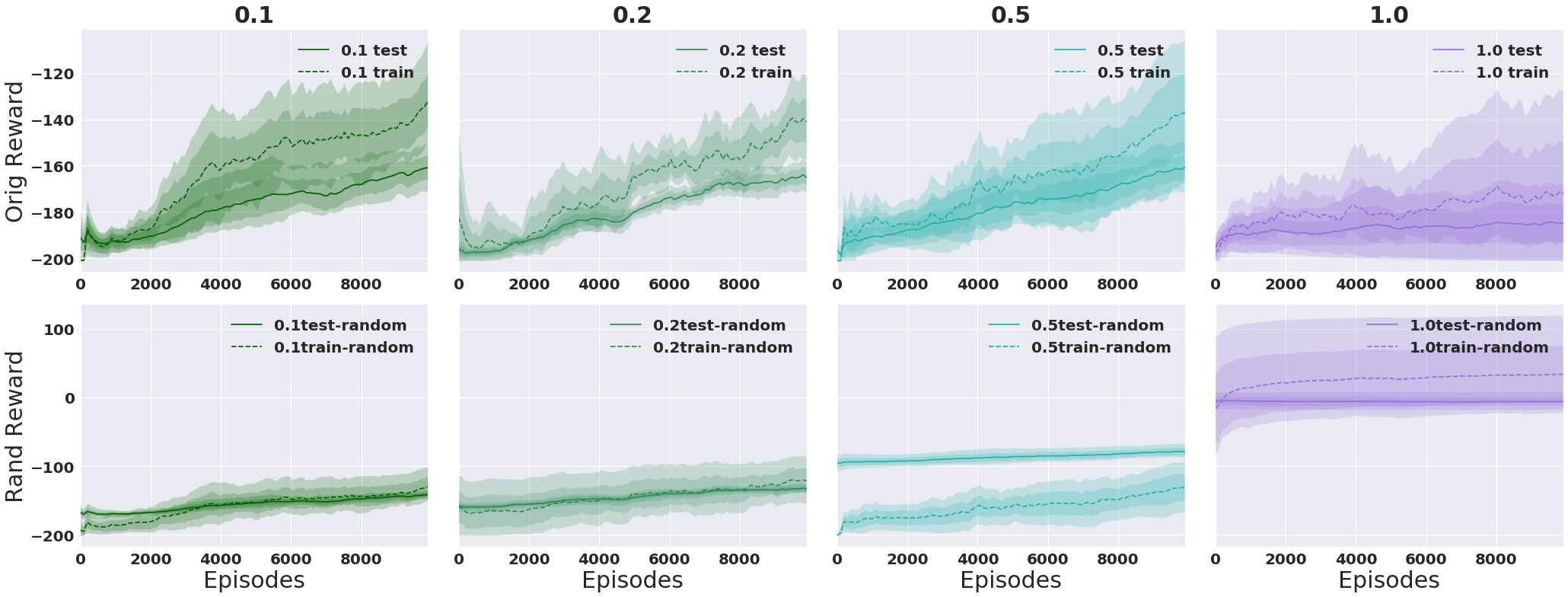}
\includegraphics[trim={0 0cm 0 0cm},clip,width=1\linewidth]{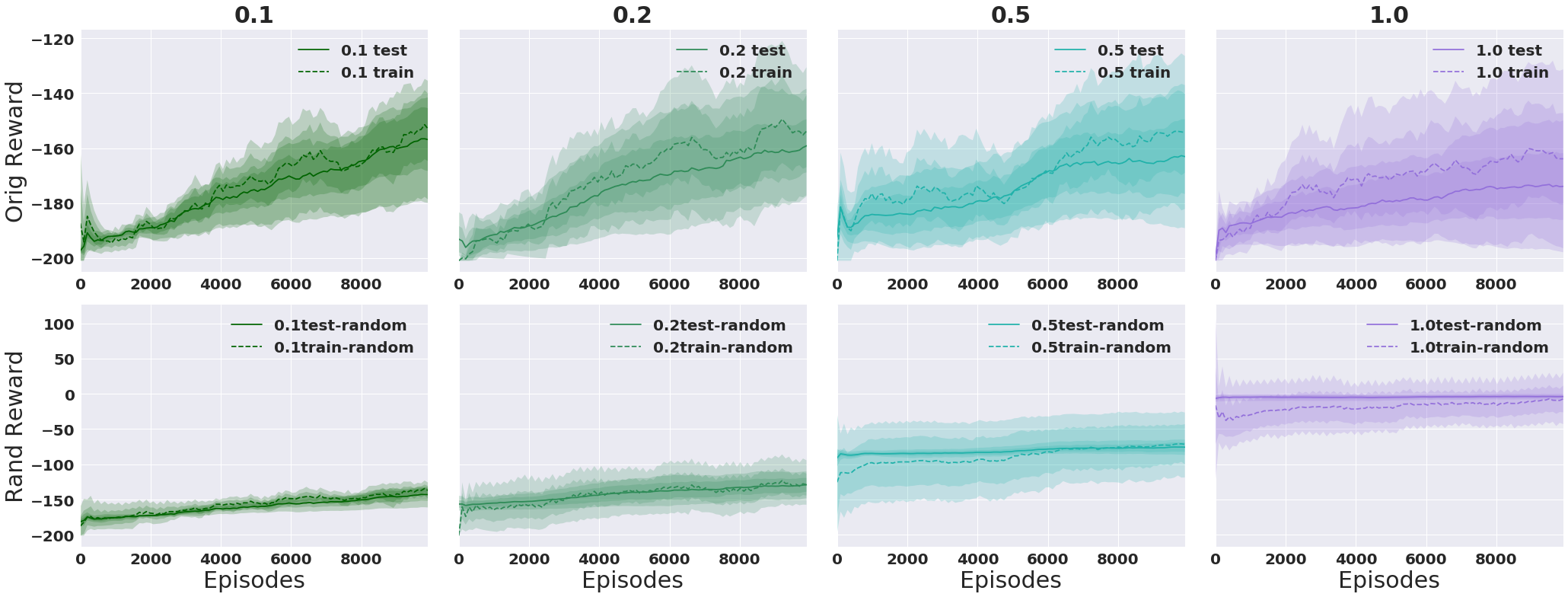}
\includegraphics[trim={0 0cm 0 0cm},clip,width=1\linewidth]{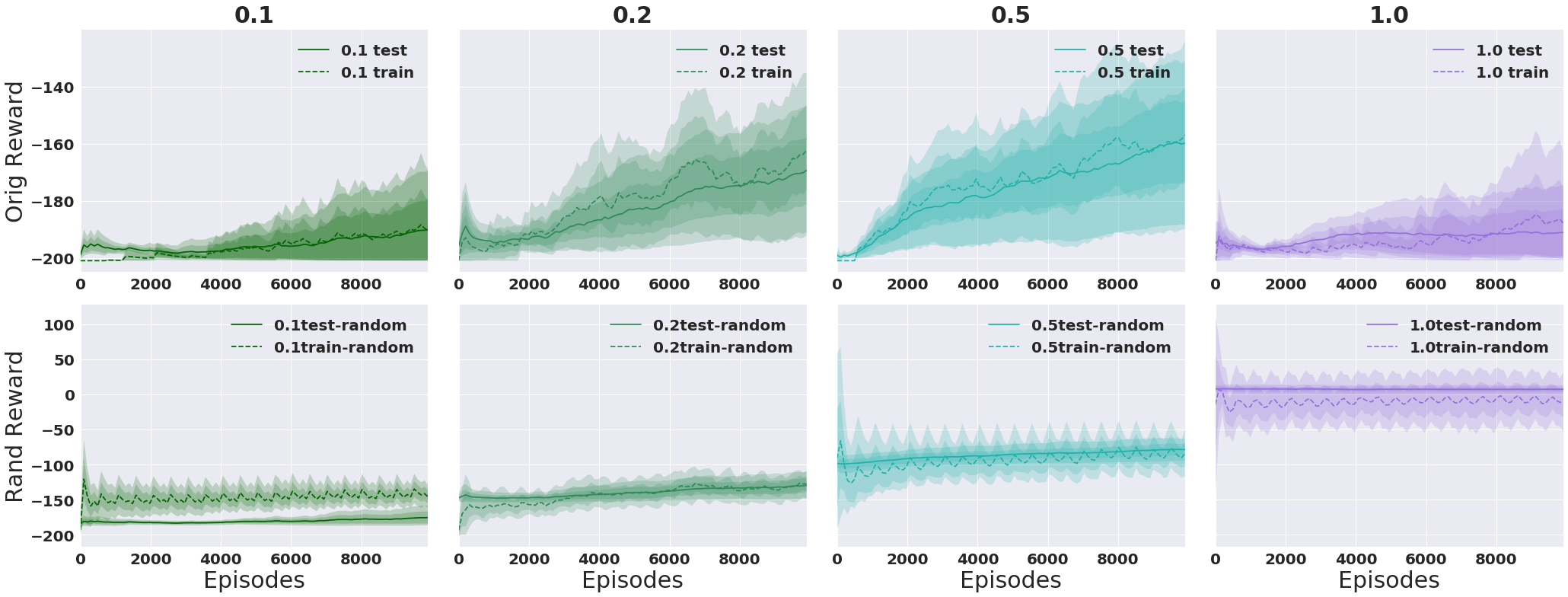}
\includegraphics[trim={0 0cm 0 0cm},clip,width=1\linewidth]{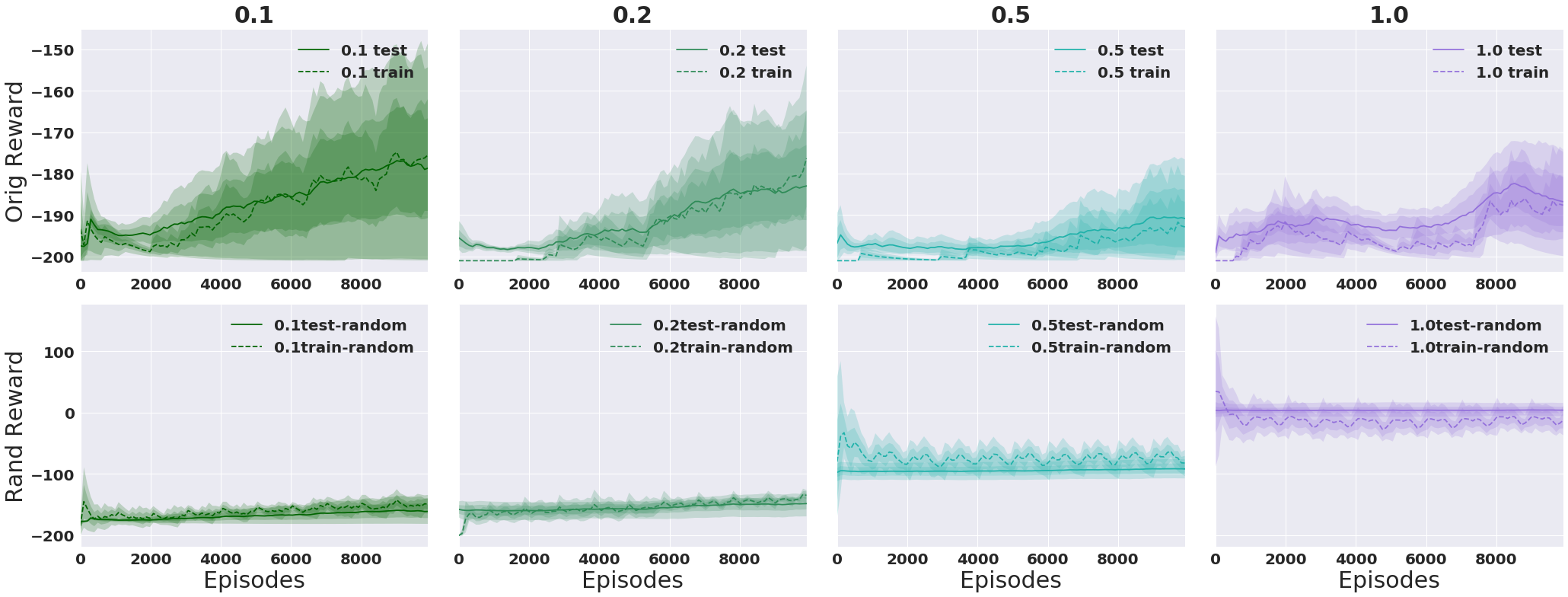}
 \caption{$b=3$. Full experiments on randomizing reward on Acrobot: Varying $p$ (cols) with $\gamma=0.99$ and varying 1, 2, 5 and 10 training seeds (rows). Averaged over 5 runs. 100K episodes.}
\label{fig:rand_bin3_acrobot}
\end{figure}

 \begin{figure}[h]
\includegraphics[trim={0 0cm 0 0cm},clip,width=1\linewidth]{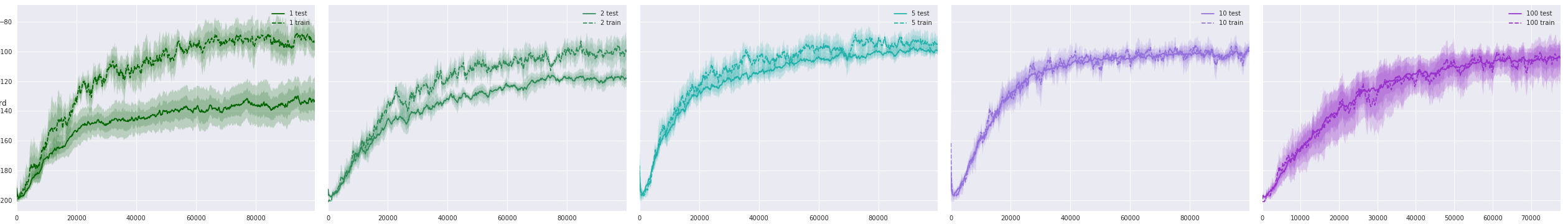}
 \caption{Reward for Acrobot varying number of training seeds. Averaged over 5 runs.}
\label{fig:acrobot_num_seeds_full_r}
\end{figure}

\subsection{Reacher}
\label{app:reacher}
Reward on a log scale. (negated, log, then negated again) \\
Number of frames: 10M

% \begin{figure}
% \centering
% \includegraphics[trim={0 0 0cm 0},clip,width=.45\linewidth]{figs/Reacher-v2_num_seeds_modelbased_gamma099_lr00003_numsteps2048_procs16_numseeds1.png}
%  \caption{Model-based. Reacher. Varying number of train seeds with $\gamma=0.99$. Averaged over 5 runs.}
% \label{fig:reacher_num_seeds_modelbased}
% \end{figure}

\subsubsection{Random Reward}
For Reacher, we randomize reward by binning $\theta$, the angle of the first link. 
 \begin{figure}
\includegraphics[trim={0 0cm 0 0cm},clip,width=1\linewidth]{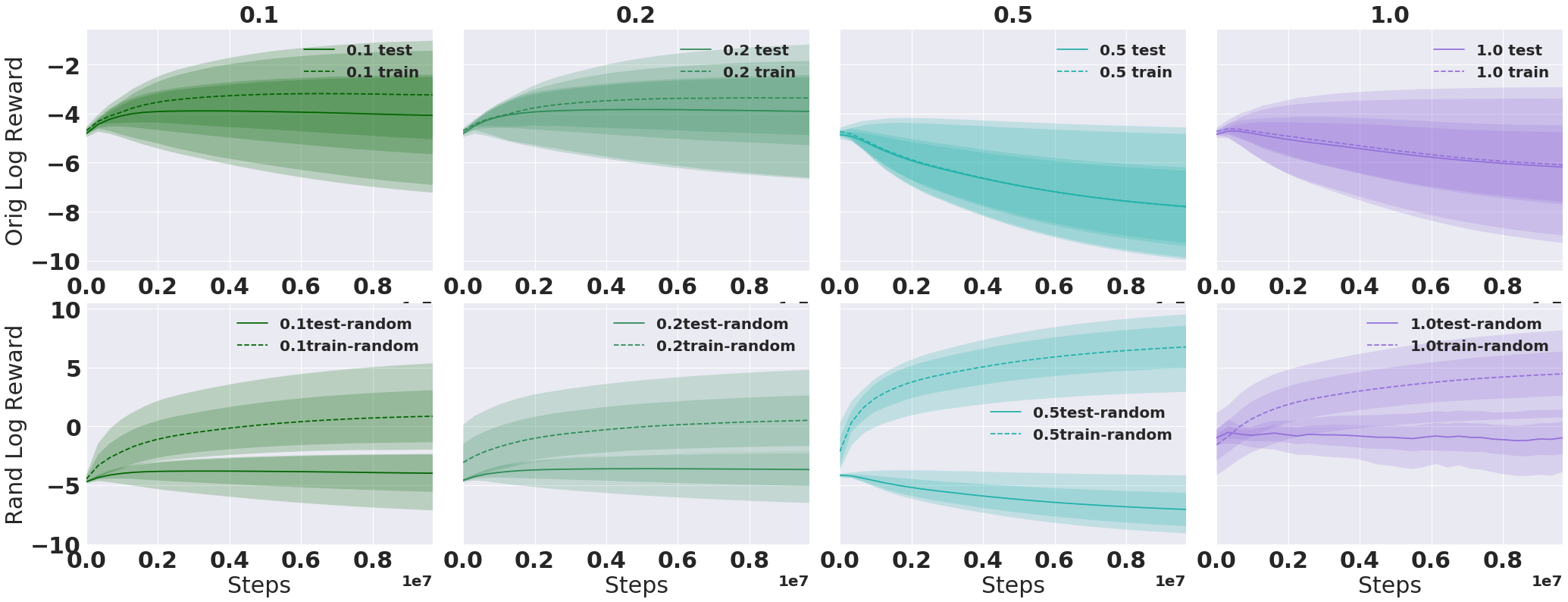}
\includegraphics[trim={0 0cm 0 0cm},clip,width=1\linewidth]{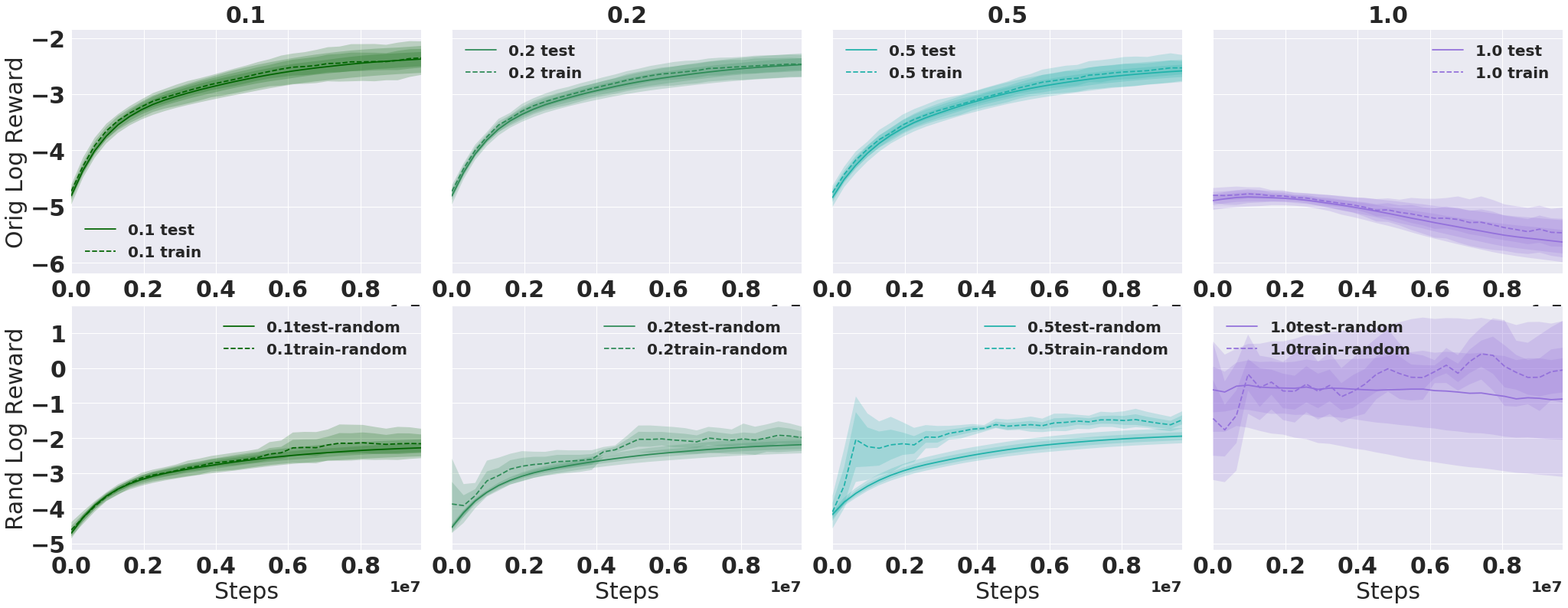}
 \caption{$b=3$. Full experiments on randomizing reward on Reacher: Varying $p$ (cols) with $\gamma=0.99$ and 1 and 100 training seed (rows). Averaged over 5 runs.}
\label{fig:rand_bin3_reacher}
\end{figure}

 \begin{figure}
\includegraphics[trim={0 0cm 0 0cm},clip,width=1\linewidth]{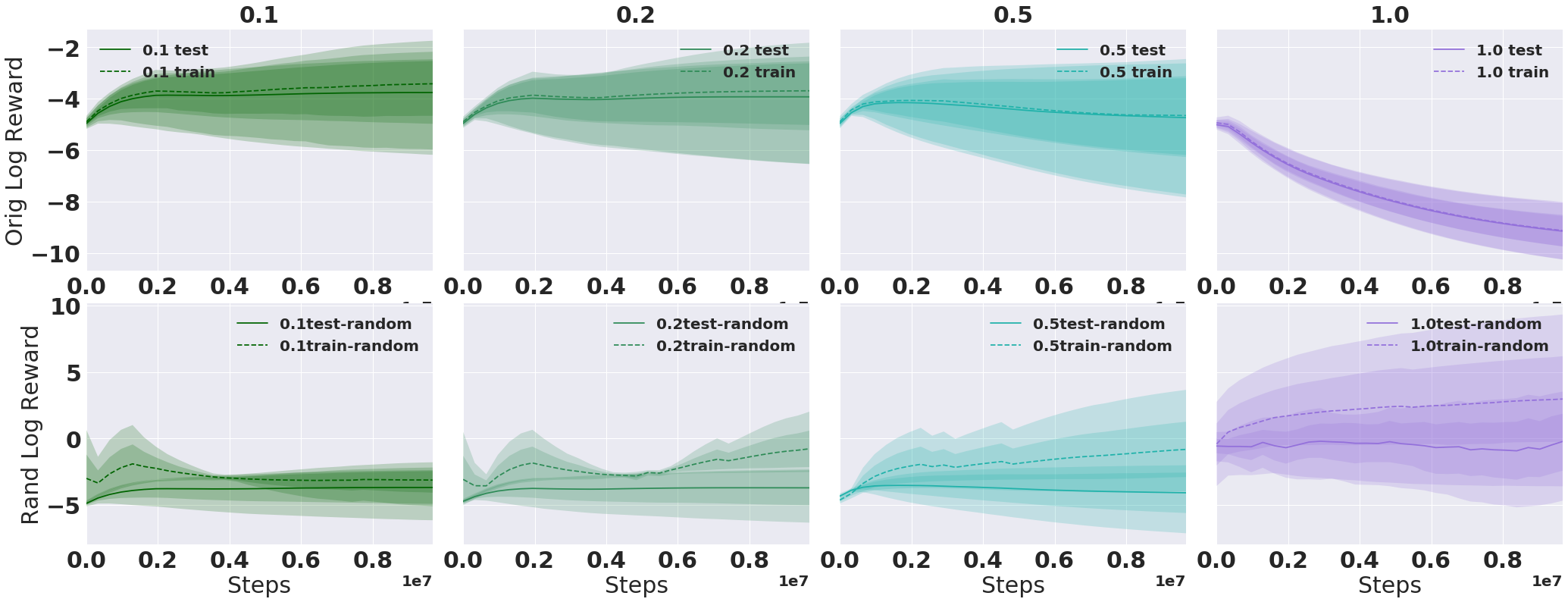}
\includegraphics[trim={0 0cm 0 0cm},clip,width=1\linewidth]{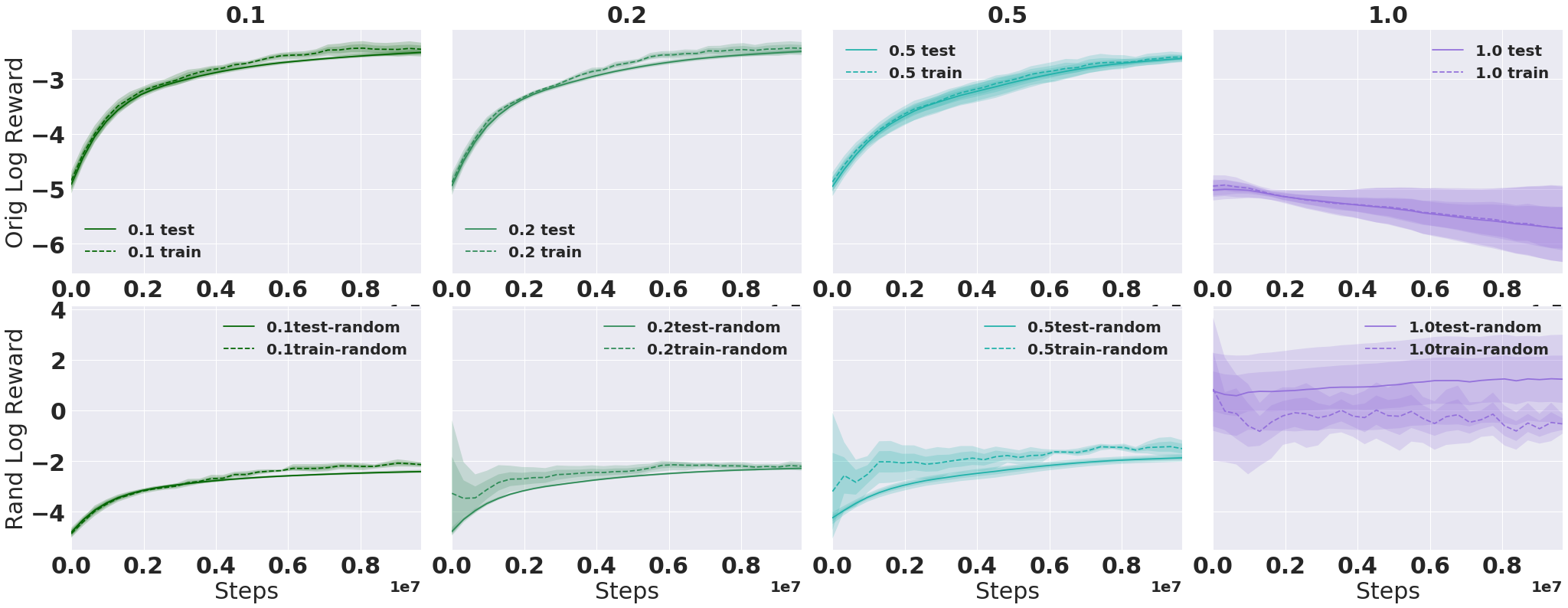}
 \caption{$b=3$. Full experiments on randomizing reward on model-based Reacher: Varying $p$ (cols) with $\gamma=0.99$ and 1 and 100 training seeds (rows). Averaged over 5 runs.}
\label{fig:rand_bin3_reacher_modelbased}
\end{figure}

\subsection{Thrower}
\label{app:thrower}
Number of frames: 100M
\subsubsection{Random Reward}
We bin the first dimension of the observation space.

 \begin{figure}
\includegraphics[trim={0 0cm 0 0cm},clip,width=1\linewidth]{figs/Thrower-v2_random_bins3_gamma099_lr00003_numsteps2048_procs16_numseeds1.png}
\includegraphics[trim={0 0cm 0 0cm},clip,width=1\linewidth]{figs/Thrower-v2_random_bins3_gamma099_lr00003_numsteps2048_procs16_numseeds100.png}
 \caption{$k=3$. Full experiments on randomizing reward on Thrower: Varying $p$ (cols) with $\gamma=0.99$ and 1 and 100 training seeds. Averaged over 5 runs.}
\label{fig:rand_bin3_thrower}
\end{figure}

 \begin{figure}
\includegraphics[trim={0 0cm 0 0cm},clip,width=1\linewidth]{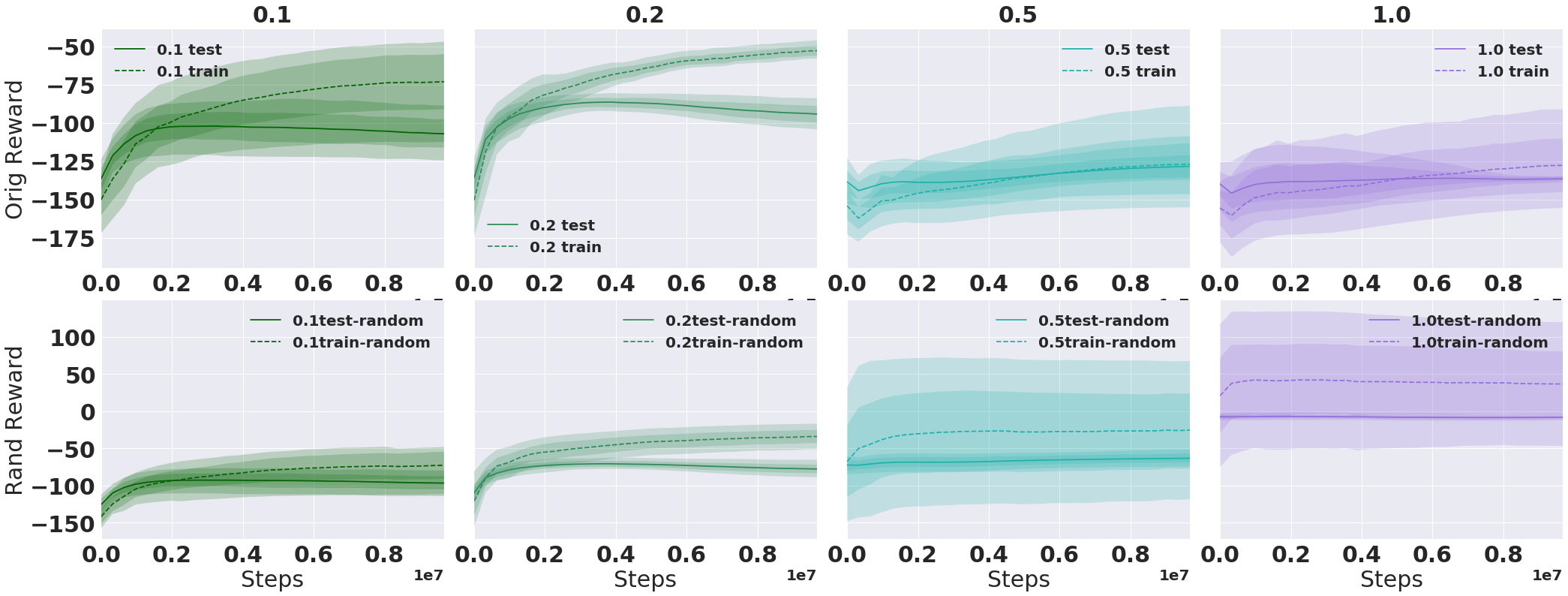}
\includegraphics[trim={0 0cm 0 0cm},clip,width=1\linewidth]{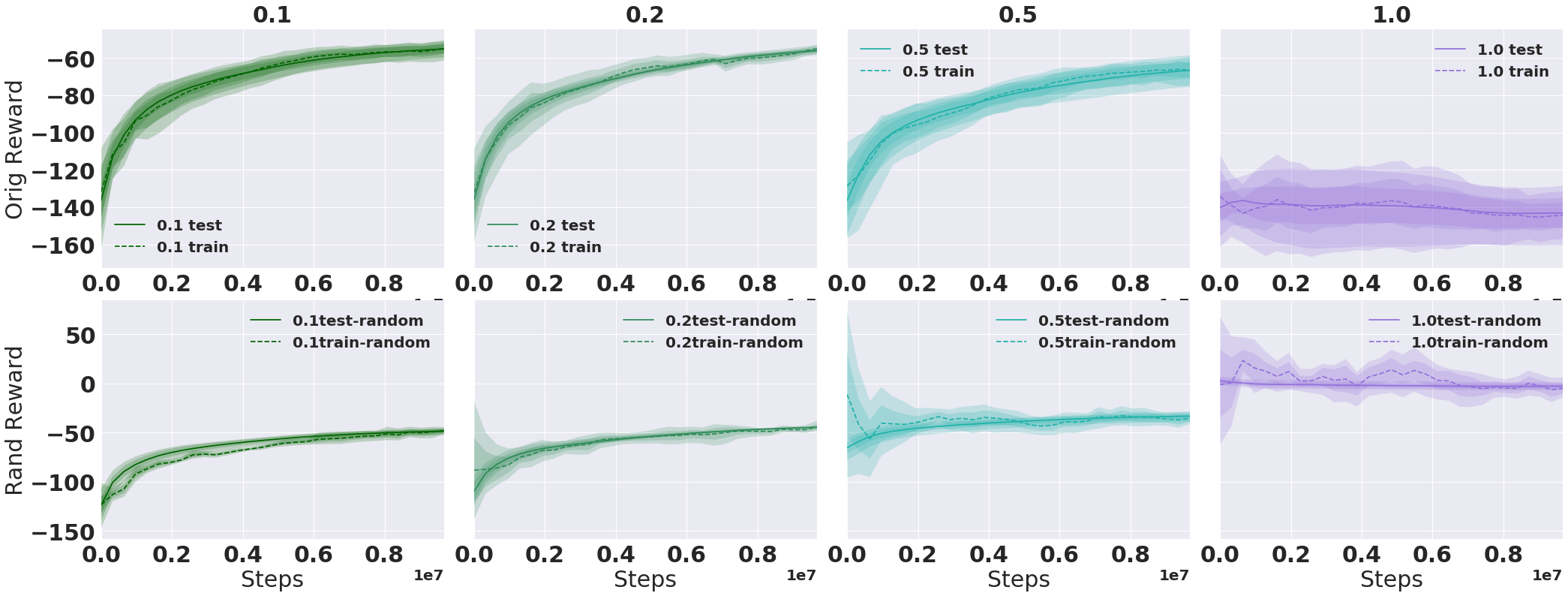}
 \caption{$b=3$. Full experiments on randomizing reward on model-based Thrower: Varying $p$ (cols) with $\gamma=0.99$ and 1 and 100 training seeds (rows). Averaged over 5 runs.}
\label{fig:rand_bin3_thrower_modelbased}
\end{figure}

\section{Transfer}
\label{app:transfer}
\begin{table}[h]
\vskip -0.1in
\begin{center}
\begin{small}
\begin{sc}
\begin{tabular}{lc|cccccc}
\hline
Environment &    Seed & $\sigma^2=0.$ & $\sigma^2=0.0001$ & $\sigma^2=0.0005$ & $\sigma^2=0.001$ & $\sigma^2=0.002$ \\
\hline
 &              1 & 	89.7  & 90.3 & 75.5  & 47.4 & 26.7\\
Cartpole & 		2 & 	131.3 & 131.3 & 112.9 &  69.5 &  35.4 \\	
& 				5 &	    105.7 & 102.7  &  66.6  & 39.1  &  19.2 \\
& 				10 &    145.5 & 144.1 & 122.8 & 101.5 &  78.8 \\
&				100 &   122.4 & 119.9  & 99.7 &  66.8 &  35.8\\
\hline
 &     & $\sigma^2=0.$ & $\sigma^2=0.1$ & $\sigma^2=0.2$ & $\sigma^2=0.3$ & $\sigma^2=0.4$ \\
\hline
             &      1 & 	19.4 & 19.2 & 18.8 & 18.9 & 19.6\\
Pixel Cartpole& 	2 & 	20.3  & 18.2 & 17.9 & 18.3 & 19.1 \\	
& 			    	5 &	    22.5 & 20.0 & 20.2 & 20.3 & 20.4 \\
& 			    	10 &    16.4 & 19.8 & 20.6 & 20.6 & 20.2\\
&			    	100 &   21.7 & 17.4 & 17.3  & 17.4 & 18.0\\
\hline
 &     & $\sigma^2=0.$ & $\sigma^2=0.001$ & $\sigma^2=0.002$ & $\sigma^2=0.003$ & $\sigma^2=0.004$ \\
\hline
 &                  1 &     31.4 & 34.2 & 17.9 & 16.3 & 15.2\\
Pixel Cartpole & 	2 & 	29.5 & 31.5 & 17.2 & 17.2 & 16.5 \\	
(Model based)& 		5 &	    35.2 & 38.2 & 20.0 & 21.7 & 21.3 \\
& 			    	10 &    31.1 & 35.7 & 19.7 & 18.6 & 18.4 \\
&			    	100 &   25.2 & 30.5 & 18.9 & 18.5 & 19.0\\
\hline
\end{tabular}
\end{sc}
\end{small}
\end{center}
%\vskip -0.1in
\caption{Transfer experiments on Cartpole and Pixel Cartpole for model-free and model-based. Columns indicate the variance $\sigma^2$ of the Gaussian noise added to the observation space. Note the difference in $\sigma^2$ of noise for all three settings.}
\vskip -0.1in
\label{table:transfer_noise_pixelcartpole}
\end{table}

 \begin{table}[h]
\vskip -0.1in
\begin{center}
\begin{small}
\begin{sc}
\begin{tabular}{lc|cccccc}
\hline
Environment &    Seed & $\sigma^2=0.$ & $\sigma^2=0.01$ & $\sigma^2=0.05$ & $\sigma^2=0.1$ & $\sigma^2=0.2$ \\
\hline
Acrobot &       1 & -130.4 & -132.1 & -156.8 & -185.7 & -198.6 \\
& 				2 & -118.7 & -120.0 & -170.8 & -193.1 & -199.0\\	
& 				5 &	-95.0 & -98.8 & -137.2 & -174.5 & -193.1 \\
& 				10 & -100.6 & -100.4 & -134.9 & -172.6 & -196.2 \\
&				100 & -107.5 &-111.9 & -155.1 & -182.8 & -196.8\\
\hline
&   &  $\sigma^2=0.$ & $\sigma^2=0.001$ & $\sigma^2=0.002$ & $\sigma^2=0.003$ & $\sigma^2=0.004$ \\
\hline
&               1 & -128.1 & -129.6 & -198.9 &  -199.9 &  -200.0 \\
Pixel Acrobot & 2 & -109.8 & -109.2 & -199.9 & -200.7 &  -200.7\\	
& 				5 &	 -87.2 & -89.0 & -200.2 & -199.8 & -200.2\\
& 				10 & -96.8 & -97.6 & -199.9 & -199.6 & -200.2\\
&				100 & -83.6 & -84.7 & -200.3 & -199.9 & -200.7\\

\end{tabular}
\end{sc}
\end{small}
\end{center}
%\vskip -0.1in
\caption{Transfer experiments on Acrobot. Columns indicate the variance $\sigma^2$ of the Gaussian noise added to the observation space.}
\vskip -0.1in
\label{table:transfer_noise_acrobot}
\end{table}

\begin{table}[hbt!]
\vskip -0.1in
\begin{center}
\begin{small}
\begin{sc}
\begin{tabular}{lc|cccccc}
\hline
Environment &    Seed & $\sigma^2=0.$ & $\sigma^2=0.1$ & $\sigma^2=0.2$ & $\sigma^2=0.5$ & $\sigma^2=1.$ \\
\hline
Reacher &       1 & 	-11.3 & -11.8 & -12.1 & -13.5 & -15.2 \\
& 				2 & 	-27.2 &  -27.1 & -26.1 & -24. & -24.6\\	
& 				5 &	    -13.8 & -14.9 & -15.5 & -16.7 & -18.8 \\
& 				10 &    -11.8 & -11.7 & -12.1 & -13.8 & -13.7 \\
&				100 &   -11.4 & -11.4 & -11.7 & -12.7 & -14.3\\
\hline
\end{tabular}
\end{sc}
\end{small}
\end{center}
%\vskip -0.1in
\caption{Transfer experiments on Reacher. Columns indicate the variance $\sigma^2$ of the Gaussian noise added to the observation space.}
\vskip -0.1in
\label{table:transfer_noise_reacher}
\end{table}

\end{document}